\documentclass[journal]{IEEEtran}
\usepackage[table]{xcolor}
\newcommand{\yb}{\cellcolor{gray!30}}
\usepackage[normalem]{ulem}
\usepackage{soul}

\usepackage{multirow}
\usepackage{subfigure}
\usepackage{url}
\usepackage{comment}
\usepackage{tikz}
\usepackage{tikz-qtree}
\usepackage{graphicx}  
\usepackage{epstopdf}
\usepackage{amsmath}
\usepackage[noadjust]{cite}

\newcommand{\sect}{Section~}
\begin{document}

\title{Lightweight and Unobtrusive Data Obfuscation at IoT Edge for Remote Inference}

\author{Dixing~Xu,~\IEEEmembership{Student~Member,~IEEE,}
	Mengyao~Zheng,
	Linshan~Jiang,~\IEEEmembership{Student~Member,~IEEE,}
	Chaojie~Gu,~\IEEEmembership{Student~Member,~IEEE,}
	Rui~Tan,~\IEEEmembership{Senior~Member,~IEEE,}
        and~Peng~Cheng,~\IEEEmembership{Member,~IEEE}
        \thanks{This work was supported in part by an NTU Start-up Grant, in part by an MOE AcRF Tier 1 grant (2019-T1-001-044), in part by HP-NTU Digital Manufacturing Corporate Lab (AI-003) funded by the Singapore Government through the Industry Alignment Fund-Industry Collaboration Projects Grant, in part by NSFC under grants 61761136012 and 61533015.}
	\thanks{Dixing Xu and Mengyao Zheng contributed equally to this research. They are with Xi'an Jiaotong-Liverpool University. This work was completed when Dixing Xu was visiting Nanyang Technological University (NTU) and then Zhejiang University (ZJU), and when Mengyao Zheng was visiting NTU. (e-mail: \{dixing.xu15, mengyao.zheng16\}@student.xjtlu.edu.cn)}
	\thanks{Linshan Jiang, Chaojie Gu and Rui Tan are with NTU. (e-mail: \{linshan001, gucj, tanrui\}@ntu.edu.sg)}
        \thanks{Peng Cheng is with ZJU. (e-mail: pcheng@iipc.zju.edu.cn)}
		\thanks{Copyright (c) 20xx IEEE. Personal use of this material is permitted. However, permission to use this material for any other purposes must be obtained from the IEEE by sending a request to pubs-permissions@ieee.org.}
}

\maketitle

\begin{abstract}
Executing deep neural networks for inference on the server-class or cloud backend based on data generated at the edge of Internet of Things is desirable due primarily to the limited compute power of edge devices and the need to protect the confidentiality of the inference neural networks. However, such a remote inference scheme incurs concerns regarding the privacy of the inference data transmitted by the edge devices to the curious backend. This paper presents a lightweight and unobtrusive approach to obfuscate the inference data at the edge devices. It is lightweight in that the edge device only needs to execute a small-scale neural network; it is unobtrusive in that the edge device does not need to indicate whether obfuscation is applied. Extensive evaluation by three case studies of free spoken digit recognition, handwritten digit recognition, and American sign language recognition shows that our approach effectively protects the confidentiality of the raw forms of the inference data while effectively preserving the backend's inference accuracy.
\end{abstract}


%
\IEEEpeerreviewmaketitle


\section{Introduction}
\label{sec:intro}

\IEEEPARstart{T}{he} fast development of sensing and communication technologies and the wide deployment of Internet-enabled smart objects in the physical environments foster the forming of the Internet of Things (IoT) as a main data generation infrastructure in the world. The tremendous amount of IoT data provides great opportunities for various applications powered by advanced machine learning (ML) technologies.

IoT in nature is a distributed system consisting of nodes equipped with sensing, computing, and communication capabilities. In order to build scalable and efficient applications on top of IoT, {\em edge computing} is a promising hierarchical system paradigm \cite{shi2016edge}. In edge computing, the widespread network edge devices (e.g., home gateways, set-top boxes, and personal smartphones) collect and process the data from the end devices that are normally smart objects deeply embedded in the physical environments (e.g., smart toothbrushes, smart body scales, smart wearables, and various embedded sensors). Then, the edge devices interact with the cloud backends of the applications to exchange data and/or commands.

Compared with the conventional cloud computing scheme that performs most of the computation in the centralized cloud servers, edge computing offers several merits including increased scalability, shortened end-to-end latency, and reduced communication bandwidth usage \cite{shi2016edge}. Deploying ML technologies (in particular, deep neural networks) in edge computing systems has attracted increasing research interests \cite{chen2019deep}. In many cases, as deep neural network designing and training are expertise-intensive and resource-consuming, IoT applications often prefer to use pre-trained inference models on the data generated at the IoT edge.
However, the implementation of the IoT edge that can leverage the latest ML technologies for inference faces two challenges:

\vspace{0.2em}
\noindent {\bf Separation of data sources and ML compute power:} With the advances of deep learning, the depth of inference models and the needed compute power to support these deep inference models increase drastically. Thus, the execution of these deep inference models on the IoT end or edge devices that have limited compute resources may be infeasible or cause too long inference time. Moreover, the execution of deep inference models on battery-based edge devices (e.g., smartphones) may not be desirable due to high power consumption. A remote server-class or cloud backend with abundant ML compute power including powerful hardware acceleration is still desired for deep inference model execution.

\vspace{0.2em}
\noindent {\bf Confidentiality of inference models:} A deployable inference model often requires significant efforts in model training and manual tuning. Thus, an inference model in general contains intellectual properties under the enterprise settings. Even when the edge devices can execute the model and meet timing/energy constraints, deploying the inference model to the edge devices in the wild may lead to the risk of intellectual property infringement (e.g., extraction of the model from the edge device memory). Moreover, the leak of the inference model can aggravate the cybersecurity concern of adversarial examples \cite{goodfellow2014explaining}. Therefore, it is desirable to protect the confidentiality of the deep inference models.

To address the above two issues, {\em remote inference} is a natural solution, in which an edge device sends the inference data to the backend, then the backend executes the inference model and sends back the result. It forms a specific interaction paradigm between the IoT edge and the backend in edge computing systems. There are existing applications adopting remote inference. PictureThis \cite{PictureThis}, a mobile App, captures a picture of plant using the smartphone's camera and then sends the picture to the cloud backend that runs an inference model to identify the plant. Amazon Alexa, a voice assistant, processes captured voices locally and also transmits the voice recordings to the cloud backend for further analysis and storage \cite{alexa-spy}. However, remote inference in the context of edge computing inevitably incurs privacy concerns, especially when the inference data at the IoT edge is collected in the user's private space and time, such as voice recordings in households \cite{alexa-spy}. The pictures for plant recognition may also be misused by the curious cloud backend to infer the users' locations based on the background of the pictures. In particular, the lack of privacy protection in remote inference may go against the recent legislation such as the General Data Protection Regulation in European Union.

Therefore, privacy preservation mechanisms are needed for remote inference in edge computing. To this end, CryptoNets \cite{gilad2016cryptonets} has been proposed to homomorphically encrypt the inference data, perform inference based on the encrypted data, and generate encrypted results. While CryptoNets provides a strong protection of the confidentiality of the inference data, it incurs significant compute overhead to the edge devices \cite{jiang2019lightweight}. Specifically, the homomorphic encryption of a $28 \times 28$ grayscale image takes about ten minutes on a Raspberry Pi 2 Model B single-board computer that has a $900\,\text{MHz}$ quad-core ARM processor. Differently, in this paper, we aim to design a {\em lightweight} data obfuscation approach suitable for resource-constrained edge devices to protect inference data privacy in the remote inference scheme, when the inference model at the backend is a pre-trained deep neural network. With the lightweight approach, the edge device spends little time and energy to obfuscate the inference data before transmitting to the backend. Moreover, we aim to achieve another feature of {\em unobtrusiveness}, in that i) the inference model at the backend admits both original and obfuscated inference data, and ii) the edge device does not need to indicate whether obfuscation is applied. The unobtrusiveness feature provides three advantages. First, the system is back-compatible with old edge devices that cannot be upgraded to perform the data obfuscation. Second, the edge device can easily choose to opt into or out of data obfuscation given its run-time computation and battery lifetime statuses. Third, the exemption of obfuscation indication helps improve privacy protection.

In this paper, we present {\em ObfNet}, an approach to realize the lightweight and unobtrusive data obfuscation at the IoT edge for remote inference. ObfNet is a small-scale neural network that can run at resource-constrained edge devices and introduces light compute overhead. ObfNet's sophisticated, many-to-one non-linear mapping from the input vector to the output vector offers a form of data obfuscation that can well protect the confidentiality of the raw forms of the input data. To achieve unobtrusiveness, we design a training procedure for ObfNet as follows. We assume that the backend has an in-service deep inference model (referred to as {\em InfNet}). The backend concatenates an untrained ObfNet with the InfNet and then trains the concatenated model using the training dataset that was used to train InfNet. During the training, only the weights of ObfNet are updated by backpropagation until convergence. The backend repeats the above procedure to generate sets of distinct ObfNets and transmits a unique set to each of the edge devices. Then, each edge device chooses an ObfNet randomly and dynamically from the received set and uses it for obfuscating the data for remote inference.

We evaluate the ObfNet approach by three case studies of 1) free spoken digit (FSD) recognition, 2) MNIST handwritten digit recognition, and 3) American sign language (ASL) recognition. The case studies show the effectiveness of ObfNet in protecting the confidentiality of the raw forms of the inference data while preserving the accuracy of the remote inference. Specifically, the obfuscated samples are unrecognizable auditorily by invited volunteers for FSD and visually for MNIST and ASL, while the obfuscation causes inference accuracy drops of generally within 1\% from the original inference accuracy of about 99\%. We also benchmark the ObfNet approach on a testbed consisting of i) a Coral development board equipped with Google's edge tensor processing unit (TPU) that acts as an edge device and ii) an NVIDIA Jetson AGX Xavier equipped with a Volta graphics processing unit (GPU) that acts as the backend. Measurements on the testbed show that i) the energy expenditures of executing ObfNet at the edge for the three case studies is at most $10\,\text{mJ}$ per sample, ii) the per-sample ObfNet execution time on the edge device is just a few milliseconds, and iii) remote inference in edge computing is advantageous in terms of total processing times and energy expenditures.

The reminder of this paper is organized as follows. \sect\ref{Related Work} reviews related work. \sect\ref{section3} states the problem and overviews our approach. \sect\ref{sec:case_studies} presents performance evaluation via three case studies. \sect\ref{sec:implementation} presents benchmark results on the testbed. \sect\ref{sec:conclusion} concludes this paper.

\section{Related Work}
\label{Related Work}

Edge computing is a new computing scheme that moves the computation and storage from the centralized cloud servers to the network edge nodes in the proximity of the end devices. It is based on a hierarchical architecture of end devices, edge nodes, and cloud servers. It brings two major advantages compared with the cloud computing scheme \cite{shi2016edge}: First, the physical proximity of the edge nodes to the end devices can reduce the end-to-end latencies if the data processing can be accomplished at the edge nodes. Second, if the cloud servers need to be involved, the data pre-processing at edge devices can reduce the communication network bandwidth usage. ML technologies, which are considered effective analytic tools for the massive data generated by the IoT end devices, have been introduced to various edge computing applications. However, there are privacy concerns when deploying ML technologies if the applications involve privacy-sensitive data. The reference \cite{chen2019deep} proposes a brief taxonomy of privacy-preserving inference and training in the context of edge computing. However, it only discusses the methods of noisification and secure computation. In what follows, we provide a more extensive taxonomy on privacy-preserving ML in edge computing.

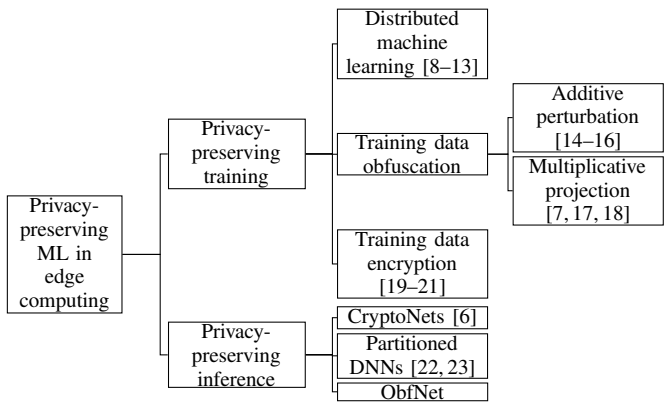
\begin{figure} 

	\usetikzlibrary{trees}
	\usetikzlibrary{patterns}
	
	\begin{tikzpicture}[grow'=right,level distance=1in]
	
	\tikzset{edge from parent/.style= 
		{ draw, edge from parent fork right},
		every tree node/.style={draw, font=\small,minimum size=0.5mm, inner sep=0.1mm, draw, minimum width=0.7in,text width=0.65in,align=center}}

	\tikzstyle{title} = [draw, font=\footnotesize,minimum size=0.1mm, inner sep=0.2mm, draw, minimum width=0.6in,text width=0.5in,align=center]
    \tikzstyle{a} = [draw, font=\footnotesize,minimum size=0.1mm, inner sep=0.2mm, draw, minimum width=0.6in,text width=0.7in,align=center]
    \tikzstyle{b} = [draw, font=\footnotesize,minimum size=0.1mm, inner sep=0.2mm, draw, minimum width=0.6in,text width=0.77in,align=center]
    
	\tikzset{level 1/.style={sibling distance=0.04in, level distance=0.9in }  }
	\tikzset{level 2/.style={sibling distance=0.02in, level distance=0.92in} }			
	\tikzset{level 3/.style={sibling distance=0.02in, level distance=0.92in} }

	\Tree 
	[.\node[title] {Privacy-preserving ML in edge computing};
	[.\node[a]
	{Privacy-\\preserving\\training};
	[.\node[b]
	{Distributed machine learning \cite{hamm2015crowd,shokri2015privacy,mcmahan2016communication,dean2012large,zinkevich2010parallelized,boyd2011distributed} };
	]  
	[.\node[b]{Training data obfuscation};
	[.\node[b]
	{Additive perturbation \cite{dwork2006calibrating,mcsherry2007mechanism,roth2010interactive}  };
	]
	[.\node[b]
	{Multiplicative projection \cite{liu2012cloud,jiang2019lightweight,shen2018privacy}  };
	]
    ]
	[.\node[b]
	{Training data encryption \cite{graepel2012ml,zhan2005privacy,qi2008efficient}};
	]
	]
	[.\node[a]
	{Privacy-\\preserving inference};
	[.\node[b]
	{CryptoNets \cite{gilad2016cryptonets} };  
	]
	[.\node[b]
	{Partitioned DNNs \cite{osia2017hybrid,wang2018not} }; 
	]
	[.\node[b]
	{ObfNet};
	]
	] 
	]
	\end{tikzpicture}
	\caption{A taxonomy of privacy-preserving ML approaches.}
	\label{Fig.main2}
	
\end{figure}

As illustrated in Fig.~\ref{Fig.main2}, existing privacy-preserving mechanisms that can be applied to the ML-equipped edge computing are categorized into privacy-preserving training and privacy-preserving inference approaches. The nodes in a privacy-preserving ML system often have two roles of {\em participant} and {\em coordinator}. When these privacy-preserving mechanisms are applied in edge computing, usually the participants are the edge nodes and the coordinator is the backend server.

In a privacy-preserving training process orchestrated by the coordinator, the participants collaboratively train a global model from their disjoint training datasets while the privacy of the training datasets is preserved. This scheme is also called {\em edge training} \cite{chen2019deep}. Distributed machine learning (DML) \cite{hamm2015crowd,shokri2015privacy,mcmahan2016communication,dean2012large,zinkevich2010parallelized,boyd2011distributed} is a typical scheme of this category, in which only model weights are exchanged among the nodes. However, the local model training and the iterative weight exchanges are compute- and communication-intensive. If the training data samples are to be transmitted to the coordinator, they can be obfuscated or encrypted for data privacy protection. Obfuscation is often achieved via additive perturbation and multiplicative projection. Additive perturbation implemented via Laplacian \cite{dwork2006calibrating}, exponential \cite{mcsherry2007mechanism}, and median \cite{roth2010interactive} mechanisms can provide differential privacy \cite{dwork2011differential}. Multiplicative projection \cite{liu2012cloud,jiang2019lightweight,shen2018privacy} protects the confidentiality of the raw forms of the original data. In \cite{liu2012cloud,jiang2019lightweight}, the participants use distinct secret projection matrices, where the Euclidean distances among the projected data samples are no longer preserved. This can degrade the performance of distance-based ML algorithms. To address this issue, in \cite{liu2012cloud}, the participants need to project a number of public data vectors and return the results to the coordinator that will learn a regress function to preserve Euclidean distances. In \cite{jiang2019lightweight}, deep neural networks (DNNs) are used to learn the sophisticated patterns of the projected data from multiple participants. ML can be also performed based on homomorphically encrypted data samples \cite{demillo1978foundations,graepel2012ml,zhan2005privacy,qi2008efficient}. However, homomorphic encryption incurs high compute overhead (millions times higher than multiplicative projection \cite{jiang2019lightweight}) and data swelling.

In privacy-preserving remote inference, the coordinator has a pre-trained inference model; the participants transmit unlabeled data samples to the coordinator for inference, while the participants' privacy in the inference data should be preserved. The proposed ObfNet is a privacy-preserving inference approach in the context of edge computing. We now review the existing privacy-preserving remote inference approaches including CryptoNets \cite{gilad2016cryptonets} and partitioned DNN approaches \cite{osia2017hybrid,wang2018not}. CryptoNets \cite{gilad2016cryptonets} adjusts the feed-forward neural network trained with plaintext data such that it can be applied to the homomorphically encrypted data to make encrypted inference. It can be used as a privacy-preserving remote inference approach in edge computing since the the coordinator will run the inference model and the edge devices encrypt the transmitted data for privacy protection. However, the high compute overhead of homomorphic encryption renders CryptoNets unpractical for edge devices. Moreover, the neural network of CrytoNets needs to use square polynomials as the activation functions, which are rare for existing neural networks that often adopt the sigmoid function or rectified linear unit (ReLU).
	
In \cite{osia2017hybrid,wang2018not}, DNN partition approaches are proposed for privacy-preserving remote inference. Specifically, a trained DNN is split into two parts. The first part, which can be considered a {\em feature extractor}, is executed by the participant, while the second part (i.e., inference model) is executed by the coordinator. For privacy protection, various alterations are applied on the feature vector extracted by the participant, which include dimension reduction and Siamese fine-tuning in \cite{osia2017hybrid}, and nullification and additive noisification for differential privacy in \cite{wang2018not}. The inference model is retrained using the altered feature vectors of the training data samples. A major limitation of the DNN partition approach \cite{wang2018not} when it is applied to edge computing is that the feature extractor at the edge nodes needs to be unique. Thus, all edge nodes need to use the same feature extractor. This renders the system vulnerable to the collusion between any single edge node and the curious backend server, because the backend may reconstruct other edge nodes' original inference data samples once they obtain the feature vector alteration mechanism. Moreover, the edge nodes cannot choose to opt out of the privacy protection, whereas our ObfNet approach allows the edge devices to choose to opt in or out freely. The feature extractor in \cite{osia2017hybrid} consists of 11 to 13 convolutional layers, which incur considerable compute overhead to edge devices.

From the above review, the training data obfuscation implemented via additive perturbation or multiplicative projection is a lightweight privacy-preserving {\em edge training} approach that can be suitable for resource-constrained IoT edge and even end devices. In contrast, lightweight privacy-preserving {\em inference} has received limited research. In particular, as IoT applications may prefer to use pre-trained deep InfNets, the development of a lightweight privacy-preserving inference approach that can adopt pre-trained InfNets is meaningful. Moreover, it is desirable if the approach introduces privacy preservation unobtrusively such that no modifications are needed for legacy edge devices and backend that were designed with no privacy preservation considerations. To achieve these goals, in this paper, we design and present ObfNet.

Our prior work \cite{zheng2019challenges} mainly focused on reviewing existing privacy-preserving machine learning schemes and discussing the challenges of applying them to IoT. It also presented the basic idea of ObfNet and preliminary results of applying ObfNet to a case study of handwritten digit recognition \cite{zheng2019challenges}. Based on \cite{zheng2019challenges}, we make the following new contributions in this paper. First, we formally define the research problem of lightweight and unobtrusive data obfuscation for remote inference in edge computing. Second, we present two new case studies of free-spoken digit recognition and American sign language recognition. A more complete set of results on the case study of handwritten digit recognition is also provided. Third, Section~\ref{sec:implementation} of this paper presents the implementation of ObfNet on a hardware testbed and the evaluation result.

\section{Problem Statement and Approach Overview}
\label{section3}

In this section, we state the privacy preservation problem in remote inference systems in the context of edge computing (\sect\ref{3.1 problem statement}) and then present the overview of the proposed ObfNet approach (\sect\ref{3.2 approach overview}).

\subsection{Problem Statement}
\label{3.1 problem statement}

We consider a remote inference system in edge computing that consists of multiple resource-constrained edge devices and a resourceful backend. The backend can be a server program in the cloud. The edge devices send the inference data samples to the backend for inference. The backend executes a pre-trained \underline{inf}erence neural \underline{net}work (InfNet) using the inference data samples. If the edge devices require the inference results, the backend sends the results to the edge devices. This remote inference scheme is advantageous if the heavyweight InfNet causes too long execution time or is not feasible on the resource-constrained edge devices.

Remote inference leads to privacy concerns if the inference data samples are privacy-sensitive. In particular, the inference data samples may contain private information beyond the inference application. Therefore, in this paper, we aim to protect the confidentiality of the raw form of each inference data sample. The data form confidentiality is an immediate and basic privacy requirement in many applications. In the experiments conducted in this paper (cf.~\sect\ref{sec:case-studies}), we use the human's ability to interpret the protected inference data samples as a measure of privacy preservation. The inference results generated by the backend may also contain information about the corresponding edge devices. However, in this paper, we do not consider the privacy contained in the inference results, since the edge devices should have no expectation of it if they are willing to join the remote inference system.

Remote inference has two major privacy threats:

\vspace{0.2em}
\noindent {\bf Honest-but-curious backend.} The backend follows the privacy preservation mechanism described in \sect\ref{3.2 approach overview} to honestly serve the edge devices. It does not intend to tamper with any data exchanged with the edge devices. However, the backend is curious about the edge devices' private information contained in the inference data, since the backend may benefit from the private information irrelevant to the objective of the inference application. For example, the backend may misuse the extracted private information for unauthorized purposes, e.g., targeted advertisement and political advocacy \cite{cambridge}.

\vspace{0.2em}
\noindent {\bf Potential collusion between edge devices and the backend.} We assume that the edge devices are not trustworthy in that they may collude with the backend in finding out other edge devices' privacy contained in the inference data. The colluding participants are also honest, i.e., they will faithfully transmit their inference data with or without obfuscation. We aim to maintain the privacy protection for an edge device when any or all other edge devices are colluding with the backend.

\subsection{Approach Overview}
\label{3.2 approach overview}

To address the privacy threats discussed in \sect\ref{3.1 problem statement}, in this paper, we propose an \underline{obf}uscation neural \underline{net}work (ObfNet) approach to obfuscate the inference data sample before being transmitted to the backend. In particular, the design of ObfNet aims to provide two properties of {\em light weight} and {\em unobtrusiveness} as discussed in \sect\ref{sec:intro}.

\begin{figure}
	\centering
	\includegraphics[width=0.65\columnwidth]{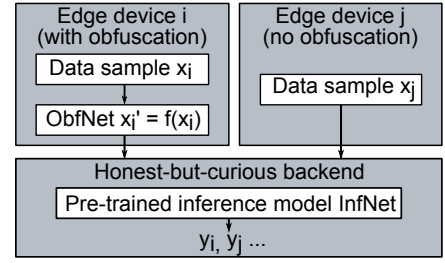}
	\caption{ObfNet for remote inference. The edge device $i$ desires privacy protection and thus applies ObfNet to obfuscate inference data sample $x_i$ to $x_i'$. The edge device $j$ does not desire privacy protection and thus directly transmits the original inference data sample $x_j$ to the backend. The backend feeds $x_i'$ and $x_j$ to the pre-trained inference model InfNet to generate the results $y_i$ and $y_j$.}
	\label{fig:voluntary}
\end{figure}

ObfNet is a small-scale neural network executed on the edge device to obfuscate the inference data samples. In our proposed approach, the backend generates multiple sets of ObfNets by following an approach detailed in the next paragraph and then transmits a unique set to each of the edge devices. An edge device that wishes to obfuscate the inference data chooses one ObfNet from the received set and feeds the inference data to the chosen ObfNet. Then, the edge device transmits the output of the ObfNet, i.e., the obfuscated inference data, to the backend for inference. The old edge devices that cannot be upgraded to perform the data obfuscation and the edge devices that do not wish to obfuscate the inference data can transmit the original inference data to the backend for inference. The backend executes the InfNet using the received inference data and sends back the inference result to the edge device. Existing cryptographic approaches can be applied to i) protect the confidentiality and integrity of the data exchanged between the edge devices and the backend and ii) the authentication of the edge devices and the backend. Fig.~\ref{fig:voluntary} illustrates the remote inference system where each edge device can choose to opt into or out of the ObfNet-based privacy preservation.

\begin{figure}
	\centering
	\includegraphics[width=1\columnwidth]{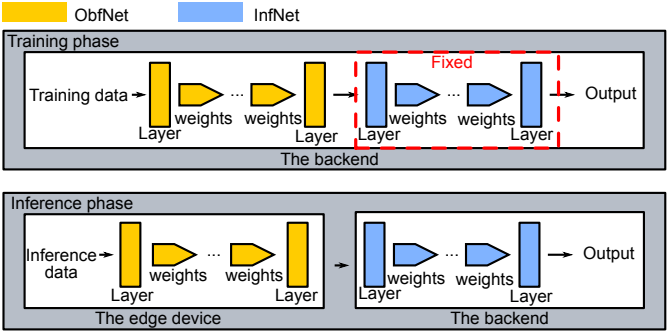}
	\caption{The procedure to generate ObfNets.}
	\label{fig:construction}
\end{figure}

Now, we present the approach to generating the sets of ObfNets at the backend. Note that the ObfNets in any set are distinct and all sets are also distinct (i.e., any two sets do not share an identical ObfNet). Fig.~\ref{fig:construction} illustrates the approach. It has two steps as follows.

\vspace{0.2em}
\noindent {\bf ObfNet design.} The system designer designs a small-scale and application-specific neural network architecture for ObfNet. The input to ObfNet is the original inference data sample. The output of ObfNet is the obfuscated inference data sample. Note that there is no rule of thumb to design ObfNet's architecture; similar to the design of DNNs for specific applications, the design of ObfNet also follows a trial-and-error approach using the validation results of the training process as the feedback (the training of ObfNet will be presented shortly). The designer should try to reduce the scale of ObfNet to make it affordable to resource-constrained edge devices. Moreover, the ObfNet design should meet the following requirements. First, to be unobtrusive, the dimensions of the input and output should be identical. Second, ObfNet should adopt many-to-one non-linear mapping activation functions (e.g., ReLU) to prevent the backend from estimating the exact original inference data from the obfuscated one.

\vspace{0.2em}
\noindent {\bf ObfNet training.} First, the backend initializes the weights of an ObfNet with random numbers. Then, the backend concatenates the ObfNet with the InfNet, forming a concatenated DNN, where the output of ObfNet is used as the input to InfNet. The backend trains the concatenated DNN using the training dataset that was previously used to train InfNet. During the backpropagation stage of each training epoch, the loss is backpropagated normally. However, only the weights of ObfNet are updated, while the weights of InfNet are fixed. When the training of the concatenated DNN converges, the backend retrieves the trained ObfNet from the concatenated DNN. By repeating the above procedure, the backend generates multiple distinct sets of distinct ObfNets. Note that due to the randomization of ObfNet's initial weights and the randomization techniques (e.g., training data sampling) during the training phase, the trained ObfNets are distinct. The backend can determine the cardinality of each set according to the available storage volume of the corresponding edge device that desires data obfuscation. Finally, the backend transmits the set to the edge device.

We have a few remarks regarding the ObfNet approach. First, since InfNet is not changed during the training of ObfNet, the InfNet can classify both the original and the obfuscated inference data samples. The execution of InfNet does not require any indication of whether the input inference data sample is obfuscated. Thus, the unobtrusive requirement is achieved. Second, as the edge devices use distinct ObfNets during remote inference, the collusion between any/all other edge devices with the backend (i.e., the colluding edge devices let the backend know which ObfNets they use) will not affect the non-colluding edge devices. Third, as the ObfNet uses many-to-one non-linear activation functions, it is highly difficult (virtually impossible) for the backend to estimate the exact original inference data sample from the obfuscated one. Moreover, as each non-colluding edge device selects an ObfNet from its received set randomly and dynamically for obfuscation, the difficulty for the backend's inverse attempt is strengthened due to the introduced uncertainty. The transmissions of the multiple ObfNets introduce a one-time overhead. In Section~\ref{sub:model-communication-overhead}, we will evaluate the amount of such overhead.

\section{Case Studies}
\label{sec:case-studies}

In this section, we present the applications of ObfNet to three case studies. For each case study, we present the data preparation, architectures of the InfNet and the ObfNet, evaluation concerning the impact of ObfNet on inference accuracy, and assessment on the quality of obfuscation. The InfNets and ObfNets are implemented using Python based on the TensorFlow library. The source code of implementation can be found from \cite{obfnet-sourcecode}.

\label{sec:case_studies}
\subsection{Case Study 1: Free Spoken Digit (FSD) Recognition}
\label{FSDD}

Our first case study concerns human voice recognition. Recently, voice recognition has been integrated into various edge systems such as smartphones and voice assistants found in households and cars. In many scenarios, voice recordings are privacy sensitive. Thus, it is desirable to obfuscate the voice data for privacy protection, while preserving the performance of voice recognition. In this section, we apply the ObfNet approach to FSD recognition, which can be viewed as a minimal voice recognition task. Using this minimal task as a case study brings the advantage of easy exposition of the results and the associated insights.

\subsubsection{Data preparation}
We use the FSD dataset \cite{free-spoken-digit-dataset:v1.0.8} that consists of 2,000 WAV recordings of spoken digits from 0 to 9 in English. We split the data as 80\% for training, 10\% for validation, and 10\% for testing. We extract the mel-frequency cepstral coefficients (MFCC) as the features to represent a segment of audio signal. MFCC is empirically shown to well represent the pertinent aspects of the short-term speech spectrum and form a particularly compact representation. As the recordings are of different lengths, we apply constant padding to unify the number of MFCC feature vectors for each recording. As a result, the extracted MFCC feature vectors over time for each recording form a $20 \times 45$ 2-dimensional image. Both the InfNet and the ObfNet take a $20 \times 45$ image as the input.

\subsubsection{Architecture of InfNet}
Multilayer perceptron (MLP) and convolutional neural network (CNN) are two types of DNNs widely adopted for speech recognition and image classification. An MLP consists of multiple fully-connected layers (or dense layers). Specifically, each neuron in any hidden layer is connected to all the neurons in the previous layer. CNN incorporates the features of shared weights, local receptive fields, and spatial subsampling to ensure shift invariance. In this case study, we design MLP-based InfNet and CNN-based InfNet, which are denoted by $I_{M}$ and $I_{C}$, respectively. Their details are as follows.

\begin{figure}
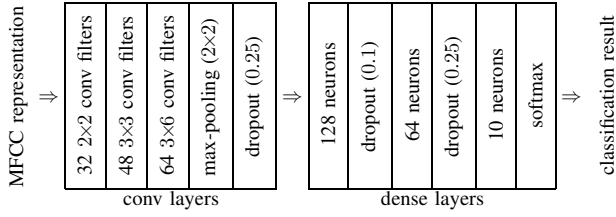

	\centering
	\normalsize
	\resizebox{0.49\textwidth}{!}{
	
	\begin{tabular}[h]{ccc|c|c|c|c|c|c|c|c|c|c|c|c|cc}
	\cline{4-8} \cline{10-15}
	& \rotatebox[origin=c]{90}{MFCC representation}  &$\!\!\!\!\Rightarrow \!\!$
	& \rotatebox[origin=c]{90}{$\;$ 32 $2\!\!\times \!\!2$ conv filters $\;$} 
	&\rotatebox[origin=c]{90}{$\;$ 48 $3\!\!\times \!\!3$ conv filters  $\;$}
	&\rotatebox[origin=c]{90}{$\;$ 64 $3\!\!\times \!\!6$ conv filters  $\;$}
	&\rotatebox[origin=c]{90}{max-pooling ($2\!\!\times \!\!2$)} 
	&\rotatebox[origin=c]{90}{dropout ($0.25$)}
	&$\!\!\Rightarrow \!\!$	 
	& \rotatebox[origin=c]{90}{128 neurons}
	&\rotatebox[origin=c]{90}{dropout ($0.1$)}
	& \rotatebox[origin=c]{90}{64 neurons}
	&\rotatebox[origin=c]{90}{dropout ($0.25$)}
	& \rotatebox[origin=c]{90}{10 neurons}
	& \rotatebox[origin=c]{90}{softmax} 
	& $\!\!\Rightarrow \!\!$  
	& \rotatebox[origin=c]{90}{classification result} 
	\\
			
	\cline{4-8} \cline{10-15}
	&\multicolumn{2}{c}{} 
	& \multicolumn{5}{c}{conv layers} 
	& \multicolumn{1}{c}{} 
	& \multicolumn{6}{c}{dense layers} 
	& \multicolumn{1}{c}{} 
	\\
	\end{tabular}
    }
	\caption{Structure of $I_C$ for FSD recognition.}
	\label{fig:Infmodel_conv}
	
\end{figure}

$I_C$ consists of three convolutional layers, one max-pooling layer, and three dense layers. %
Zero padding is performed to the input image in the convolutional layers and the max-pooling layer. %
ReLU activation is applied to the output of every convolutional and dense layer except for the last layer. ReLU rectifies a negative input to zero. %
The last dense layer has 10 neurons with a softmax activation function corresponding to the 10 classes of FSD. %
Three dropout layers with dropout rate 0.25, 0.1 and 0.25 are applied after the max-pooling layer and in the first two dense layers. %
Specifically, 25\%, 10\%, and 25\% of the neurons will be abandoned randomly from the neural network during the training process. %
Dropout is an approach to regularization in neural networks which helps reduce interdependent learning amongst the neurons. It is widely leveraged during model training to avoid overfitting. Fig.~\ref{fig:Infmodel_conv} shows the structure of $I_C$. %
The $I_C$ has about 1.1 million parameters in total. %

\begin{figure}
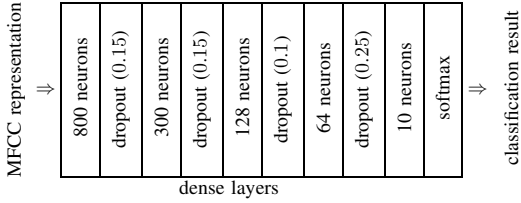

	\centering
	\resizebox{0.42\textwidth}{!}{
	\normalsize
	\begin{tabular}[h]{ccc|c|c|c|c|c|c|c|c|c|c|cc}
	
	\cline{4-13}
	& \rotatebox[origin=c]{90}{MFCC representation}  &$\!\!\!\!\Rightarrow \!\!$
	& \rotatebox[origin=c]{90}{800 neurons} 
	&\rotatebox[origin=c]{90}{dropout ($0.15$)}
	&\rotatebox[origin=c]{90}{300 neurons}
	&\rotatebox[origin=c]{90}{dropout ($0.15$)} 
	&\rotatebox[origin=c]{90}{128 neurons}
	& \rotatebox[origin=c]{90}{dropout ($0.1$)}
	& \rotatebox[origin=c]{90}{64 neurons}
	&\rotatebox[origin=c]{90}{dropout ($0.25$)}
	& \rotatebox[origin=c]{90}{10 neurons}
	& \rotatebox[origin=c]{90}{softmax} 
	& $\!\!\Rightarrow \!\!$  
	& \rotatebox[origin=c]{90}{classification result} 
	\\		
	\cline{4-13}		
	& \multicolumn{12}{c}{dense layers} 
	& \multicolumn{1}{c}{} 
	\\		
	\end{tabular}
	}
	
	\caption{Structure of $I_M$ for FSD recognition.}
	\label{fig:Infmodel_fc}
	
\end{figure}

$I_M$ has five dense layers. %
ReLU activation is applied to the output of every hidden layer.
The last dense layer has 10 neurons with a softmax activation function. %
To prevent overfitting, four dropout layers are applied after the hidden layers. %
Fig.~\ref{fig:Infmodel_fc} shows the structure of $I_M$. %
The $I_M$ has about one million parameters in total.

\subsubsection{Architecture of ObfNet}

Similar to InfNets, we design CNN-based and MLP-based ObfNets, which are denoted by $O_{C}$ and $O_M$, respectively. Their details are as follows.

$O_C$ consists of two convolutional layers, one max-pooling layer and one dense layer as the output layer. %
The first convolutional layer filters the $20 \times 45$ input image with three output filters of kernel size $2 \times 4$. %
The second convolutional layer applies five output filters with kernel size $3 \times 6$. %
All convolutional filters use a stride of one pixel. %
Batch normalization follows both convolutional layers, which is expected to mitigate the problem of internal covariate shift to improve model performance. %
A max-pooling layer with pool size of $2 \times 2$ and stride of two is then used to reduce the data dimensionality for computational efficiency. %
Zero padding is added in each convolutional layer and the max-pooling layer, to ensure that the filtered image has the same dimension as the input image in each layer. %
The dense layer with 900 neurons is then connected after flattening the output of the max-pooling layer. %
ReLU activation is applied to the output of every convolutional and dense layer. %
This introduces many-to-one mapping that is needed in our scheme as discussed in \sect\ref{3.2 approach overview}. %
Two dropout layers of with dropout rates of 0.25 and 0.15 are applied respectively after the max-pooling layer and in the dense layer.
In order to ensure that the output of ObfNet has the same size as the input, a reshape layer is applied in the end to reshape the output size to $20 \times 45$. %
The $O_C$ has about 0.65 million parameters.
	
$O_M$ has two dense layers as hidden layers. %
The first layer has 200 neurons and is fully connected to the second layer of 900 neurons. %
ReLU activation and batch normalization are applied to the output of both layers. %
A reshape layer is used as the output layer. %
The $O_M$ has about 0.37 million parameters.

\subsubsection{Inference accuracy of InfNet and ObfNet-InfNet}

Following the procedures described in \sect \ref{3.2 approach overview}, we train $I_C$ and $I_M$ using the training dataset and then train $O_C$ and $O_M$ in the four concatenations of ObfNet and InfNet (i.e., $O_C$-$I_C$, $O_M$-$I_C$, $O_C$-$I_M$, $O_M$-$I_M$). During the training phase, we adopt the AdaDelta optimizer \cite{zeiler2012adadelta}, which introduces minimal computation overhead over stochastic gradient descent (SGD) and adapts the learning rate dynamically. Note that during the training phase, only the model achieving the highest validation accuracy is yielded as the training result.

\begin{figure}
  \centering
  \includegraphics[width=.9\columnwidth]{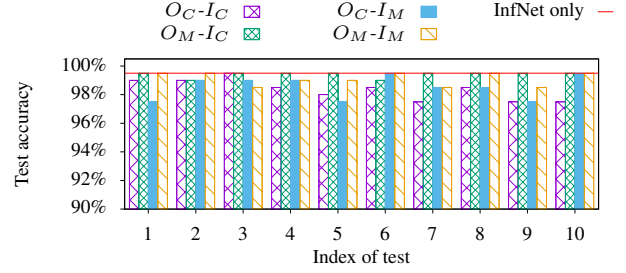}
  \caption{Test accuracy of different ObfNet-InfNet concatenations in ten tests.}
  \label{fig:fsd-acc}
\end{figure}

The test accuracy of the trained InfNets $I_C$ and $I_M$ is 99.5\%. Thus, the InfNets are well trained. %
The four ObfNet-InfNet concatenations give distinct test accuracy. For each concatenation, we trained ten different ObfNets. Fig.~\ref{fig:fsd-acc} shows the inference accuracy of applying ten different ObfNets before the well-trained InfNet. The average test accuracy of applying $O_C$ and $O_M$ before $I_C$ is 98.35\% and 99.40\%, respectively. The average test accuracy of applying $O_C$ and $O_M$ before $I_M$ is 98.55\% and 99.10\%, respectively. Compared with the test accuracy of the $I_C$ and $I_M$ on the original data (i.e., 99.5\%), the test accuracy drops caused by the obfuscation are merely 1.15\%, 0.10\%, 0.95\% and 0.40\% for different combinations of the ObfNet and the InfNet. Thus, the inference accuracy is well preserved when ObfNet is employed.

\subsubsection{Quality of obfuscation}
To understand the quality of obfuscation, we apply the MFCC inverse using a Python package LibROSA to convert the MFCC representations back to WAV audio. The audio converted from the original MFCC representations can be easily recognized by human despite some distortions. We also design an experiment to investigate whether humans can interpret the audios inverted from the outputs of ObfNet, i.e., the obfuscated MFCC representations. The details and results of the experiment are as follows.

We invited ten student volunteers (five males and five females) aged from 21 to 23 from Xi'an Jiaotong-Liverpool University. All volunteers have good hearing. In the experiment, we randomly selected ten original MFCC representations from the test dataset (one for each class of the FSD dataset). Then, we applied the MFCC inverse using LibROSA to convert the ten MFCC representations back to audio. The four different ObfNets (two $O_C$s and two $O_M$s) used in our evaluation were applied to obfuscate the two selected MFCC representations. The obfuscated MFCC representations were inverted using LibROSA to audios. Therefore, in total, there were 50 audio files: ten for the original MFCC representations and 40 for the obfuscated MFCC representations. All volunteers sat in a classroom. The 10 audio files inverted from the original MFCC representations were firstly played in the classroom in a shuffled order. All volunteers can correctly recognize the FSDs. Then, the 40 audio files inverted from the obfuscated MFCC representations were played in a shuffled order. Every volunteer was required to write down the FSD label (from 0 to 9) that they perceived.

Fig.~\ref{fig:matrix_OCIC} shows the confusion matrix for the ten volunteers to recognize the audios inverted from the MFCC representations obfuscated by ObfNet $O_C$ that is trained for InfNet $I_C$. Each row shows the distribution of the ten volunteers' answers for an audio with a certain true label. The last column shows the accuracy for the audio. From the figure, we can see that the volunteers' answers are distributed over all labels without any consensus. This suggests that the volunteers cannot perceive useful information from the audio in recognizing the FSD. The confusion matrices for the other three ObfNets can be found in Appendix~\ref{appendix: confusion-matrices}. The overall accuracy, which is defined as the number of correct answers divided by a total of 100 answers (10 volunteers $\times$ 10 audios), is 5\%, 7\%, 7\%, 4\% for the four ObfNets, respectively. Thus, the volunteers' answers seem to be random guesses with an expected accuracy of 10\%.
Therefore, the ObfNets achieve satisfactory obfuscation quality. Interested readers can download the obfuscated audio samples from an online repository \cite{ObfNet-showcase} and then examine them.

\begin{figure}
\resizebox{0.49\textwidth}{!}{
	
    \begin{tabular}{l|c|c|c|c|c|c|c|c|c|c|c|c|}
    
	\multicolumn{13}{c}{\textbf{Perceived label}}\\
	\cline{3-13}
	\multicolumn{2}{c|}{}&\textbf{0}&\textbf{1}&\textbf{2}&\textbf{3}&\textbf{4}&\textbf{5}&\textbf{6}&\textbf{7}&\textbf{8}&\textbf{9}&{Accuracy}\\
	\cline{2-13}
	\multirow{10}{*}{\rotatebox[origin=c]{90}{\textbf{True label}}}
	&\textbf{0}& \yb$ $ & $1$ & $1$&$1$&$1$&$1$&$1$&$1$&$2$&$1$&$0\%$ \\
	\cline{2-13}
	&\textbf{1}& $1$ & \yb$ $ & $1$ & $1$ & $2$ & $2$ & $1$ & $1$ & $ $ & $1$ & $0\%$\\
	\cline{2-13}
	&\textbf{2}& $ $ & $1$ & \yb$1$ & $1$ & $2$ & $ $ & $1$ & $2$ & $1$ & $1$ &$10\%$\\
	\cline{2-13}
	&\textbf{3}& $2$ & $1$ & $ $ & \yb$1$ & $1$ & $1$ & $2$ & $1$ & $ $ & $1$ &$10\%$\\
	\cline{2-13}
    &\textbf{4}& $1$ & $1$ & $1$ & $2$ & \yb$1$ & $ $ & $1$ & $2$ & $1$ & $ $ &$10\%$\\
    \cline{2-13}
    &\textbf{5}& $1$ & $ $ & $ $ & $1$ & $1$ & \yb$ $ & $1$ & $1$ & $3$ & $2$ &$0\%$\\
    \cline{2-13}
    &\textbf{6}& $1$ & $1$ & $2$ & $3$ & $1$ & $1$ & \yb$ $ & $ $ & $ $ & $1$ &$0\%$\\
    \cline{2-13}
    &\textbf{7}& $2$ & $2$ & $ $ & $2$ & $1$ & $1$ & $ $ & \yb$1$ & $1$ & $ $ &$10\%$\\
    \cline{2-13}
    &\textbf{8}& $ $ & $2$ & $1$ & $1$ & $2$ & $1$ & $1$ & $1$ & \yb$1$ & $ $ &$10\%$\\
    \cline{2-13}
    &\textbf{9}& $ $ & $1$ & $1$ & $1$ & $1$ & $1$ & $2$ & $1$ & $2$ & \yb$ $ &$0\%$\\
    \cline{2-13}
    \multicolumn{13}{c}{\textbf{Overall accuracy = 5\%}}\\

    \end{tabular}
    }

\caption{Confusion matrix for recognizing the audio inverted from the MFCC representations obfuscated by ObfNet $O_C$ that is trained for InfNet $I_C$. The matrix omits the zeros.}
\label{fig:matrix_OCIC}
\end{figure}

\subsection{Case Study 2: Handwritten Digit (MNIST) Recognition}
\label{case-study-mnist}

The MNIST dataset of handwritten digits \cite{lecun1998mnist} has been widely adopted in ML literature. In this section, we apply our ObfNet approach to MNIST. Due to the simplicity of the image samples in MNIST, the quality of the obfuscation can be readily assessed by visual inspection.

\subsubsection{Data preparation}

The MNIST dataset consists of 70,000 handwritten digit images with ten classes corresponding to the digits from 0 to 9, as shown in Fig.~\ref{fig:mnist_origin0}. Each image has a single channel (i.e., grayscale image). We resize each image to $28 \times 28$.

\subsubsection{Architecture of InfNet}

\begin{figure}[t]
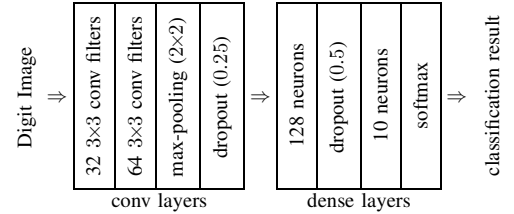

  \centering
  \normalsize
	\resizebox{0.4\textwidth}{!}{
	\begin{tabular}{ccc|c|c|c|c|c|c|c|c|c|cc}
	\cline{4-7} \cline{9-12}
	& \rotatebox[origin=c]{90}{Digit Image}  &$\!\!\!\!\Rightarrow \!\!$
	& \rotatebox[origin=c]{90}{$\;$ 32 $3\!\!\times \!\!3$ conv filters $\;$} 
	&\rotatebox[origin=c]{90}{$\;$ 64 $3\!\!\times \!\!3$ conv filters  $\;$}
	&\rotatebox[origin=c]{90}{max-pooling ($2\!\!\times \!\!2$)} 
	&\rotatebox[origin=c]{90}{dropout ($0.25$)}
	&$\!\!\Rightarrow \!\!$	 
	& \rotatebox[origin=c]{90}{128 neurons}
	&\rotatebox[origin=c]{90}{dropout ($0.5$)}
	& \rotatebox[origin=c]{90}{10 neurons}
	& \rotatebox[origin=c]{90}{softmax} 
	& $\!\!\Rightarrow \!\!$  
	& \rotatebox[origin=c]{90}{classification result} 
	\\
			
	\cline{4-7} \cline{9-12}
	&\multicolumn{2}{c}{} 
	& \multicolumn{4}{c}{conv layers} 
	& \multicolumn{1}{c}{} 
	& \multicolumn{4}{c}{dense layers} 
	& \multicolumn{1}{c}{} 
	\\
	\end{tabular}
	}
	\caption{Structure of $I_C$ for MNIST recognition.}
	\label{fig:inf-conv-mnist}
\end{figure}

\begin{figure}[t]
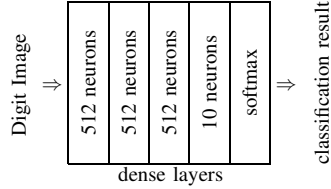

  \centering
  \normalsize
	\resizebox{0.28\textwidth}{!}{
	\begin{tabular}{ccc|c|c|c|c|c|cc}
	\cline{4-8}
	& \rotatebox[origin=c]{90}{Digit Image}  &$\!\!\!\!\Rightarrow \!\!$
	& \rotatebox[origin=c]{90}{512 neurons} 
	&\rotatebox[origin=c]{90}{512 neurons}
	&\rotatebox[origin=c]{90}{512 neurons}
	& \rotatebox[origin=c]{90}{10 neurons}
	& \rotatebox[origin=c]{90}{softmax} 
	& $\!\!\Rightarrow \!\!$  
	& \rotatebox[origin=c]{90}{classification result} 
	\\		
	\cline{4-8}		
	& \multicolumn{9}{c}{dense layers} 
	\\		
	\end{tabular}
	}
	
	\caption{Structure of $I_M$ for MNIST recognition.}
	\label{fig:inf-fc-mnist}
	
\end{figure}

We adopt two InfNets: a CNN-based $I_C$ and an MLP-based $I_M$. Their details are as follows.
$I_C$ is similar to LeNet~\cite{lecun1998mnist}. It consists of five layers: two convolutional layers, a pooling layer, and two dense layers with ReLU activation. Fig.~\ref{fig:inf-conv-mnist} shows the architecture. The $I_C$ has about 1.2 million parameters in total.
$I_M$ has four dense layers as illustrated in Fig.~\ref{fig:inf-fc-mnist}. It has about 0.93 million parameters in total.

\subsubsection{Architecture of ObfNet}
\label{subsubsec:mnist-arch-obfnet}

An MLP-based ObfNet $O_M$ and a CNN-based ObfNet $O_C$ are adopted. Details are as follows.

$O_M$ has two dense layers with ReLU activation. This two-layer design helps reduce the scale of ObfNet. Specifically, to be unobtrusive, ObfNet's output must have the same size as its input.For input size of $28 \times 28 = 784$, a single-layer MLP with bias has $784\times784+784=615440$ parameters. In contrast, a two-layer MLP with 16 neurons within each layer has $784\times16+16+16\times784+784=25888$ parameters only, which is 23.8 times smaller than the single-layer MLP. We configure the number of neurons for the first hidden layer to be $8$, $16$, $32$, $64$, or $128$. We will investigate the impact of ObfNet's scale on the accuracy of InfNet. The amounts of parameters corresponding to the above configurations are from 0.013 to 0.804 million.

$O_C$ has a convolutional layer, a pooling layer, a dropout layer, and two dense layers with ReLU activation. The convolutional layer filters the $28 \times 28$ input image with 32 output filters of kernel size $3 \times 3$ and uses stride of one pixel. The max-pooling layer with pool size of $2 \times 2$ and stride of two follows to reduce spatial dimensions. wo dense layers are then connected with ReLU activation.

\subsubsection{Inference accuracy of InfNet and ObfNet-InfNet}

\begin{figure}
\centering
	\subfigure[Test accuracy of InfNet $I_C$ and two ObfNet-InfNet concatenations]{\label{fig:mnist-obf-inf-conv}\includegraphics[width=.9\columnwidth]{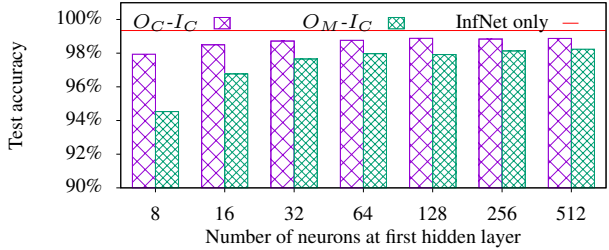}}
	\subfigure[Test accuracy of InfNet $I_M$ and two ObfNet-InfNet concatenations]{\label{fig:mnist-obf-inf-mlp}\includegraphics[width=.9\columnwidth]{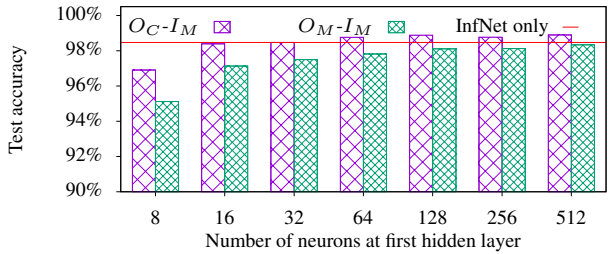}}

\caption{Test accuracy of InfNets and ObfNet-InfNet concatenations for MNIST recognition.}
\label{fig:mnist-acc-result}
\end{figure}

The test accuracies of the trained InfNets $I_C$ and $I_M$ are 99.35\% and 98.47\%, respectively. This suggests that the InfNets are well trained. As discussed in \sect\ref{subsubsec:mnist-arch-obfnet}, we vary the number of neurons of the first hidden layer of the ObfNets and train the ObfNets following the procedure presented in \sect\ref{3.2 approach overview}. Fig.~\ref{fig:mnist-acc-result} shows the test accuracy of various concatenations of ObfNets and InfNets when the number of neurons in the first hidden layer of the ObfNet varies. From Fig.~\ref{fig:mnist-obf-inf-conv}, compared with the test accuracy of $I_C$, the concatenation $O_M$-$I_C$ has test accuracy drops ranging from 0.46\% to 1.43\% over various neuron number settings. When the InfNet $I_M$ is adopted, more neurons in the first hidden layer of ObfNet result in higher test accuracy of the ObfNet-InfNet concatenation, as shown in Fig.~\ref{fig:mnist-obf-inf-mlp}. In particular, some ObfNet-InfNet concatenations even outperform the corresponding InfNet. This is possible because the ObfNet-InfNet concatenations are deeper neural networks compared with the corresponding InfNet.

\subsubsection{Quality of obfuscation}

\begin{figure}
	\centering
	\subfigure[Original inference data]{\label{fig:mnist_origin0}\includegraphics[width=1\columnwidth]{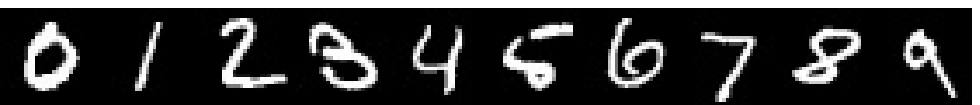}}
	\subfigure[Obfuscation results of $O_{M}$ with 8 neurons in the first hidden layer]{\label{fig:mnist_obf8}\includegraphics[width=1\columnwidth]{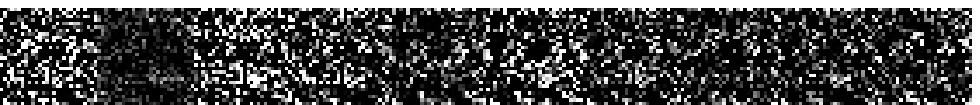}}
	\subfigure[Obfuscation results of $O_{M}$ with 16 neurons in the first hidden layer]{\label{fig:mnist_obf16}\includegraphics[width=1\columnwidth]{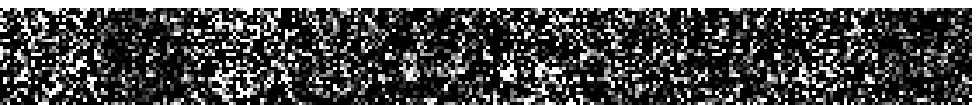}}
	\subfigure[Obfuscation results of $O_{M}$ with 32 neurons in the first hidden layer]{\label{fig:mnist_obf32}\includegraphics[width=1\columnwidth]{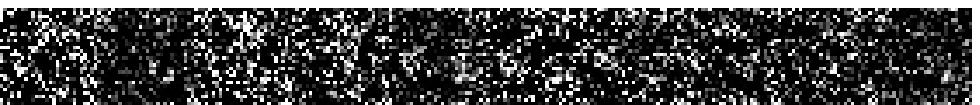}}
	\subfigure[Obfuscation results of $O_{M}$ with 64 neurons in the first hidden layer]{\label{fig:mnist_obf64}\includegraphics[width=1\columnwidth]{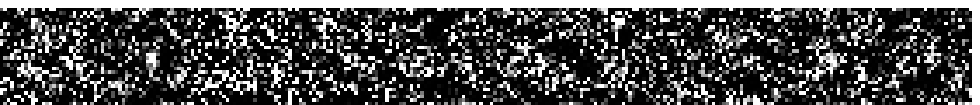}}
	\subfigure[Obfuscation results of $O_{M}$ with 128 neurons in the first hidden layer]{\label{fig:mnist_obf128}\includegraphics[width=1\columnwidth]{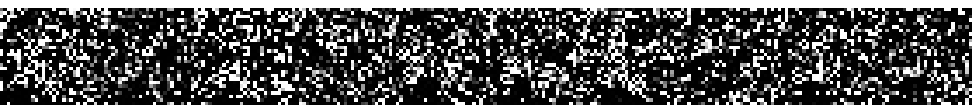}}
	\subfigure[Obfuscation results of $O_{M}$ with 256 neurons in the first hidden layer]{\label{fig:mnist_obf256}\includegraphics[width=1\columnwidth]{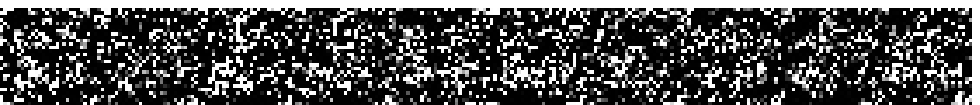}}
	\subfigure[Obfuscation results of $O_{M}$ with 512 neurons in the first hidden layer]{\label{fig:mnist_obf512}\includegraphics[width=1\columnwidth]{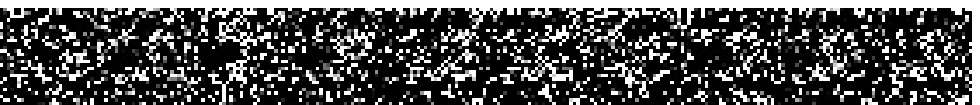}}
	\caption{Obfuscation results of ObfNet $O_M$ on MNIST.}
	\label{fig:mnist-image-result-om}
\end{figure}

Fig.~\ref{fig:mnist-image-result-om} shows the obfuscation results of $O_M$ when the number of neurons in the first hidden layer varies. From the figure, we cannot interpret the obfuscation results into any digits. When the number of neurons is few (e.g., 8 to 32), the obfuscation results of the digit one are darker than the obfuscation results of other digits. This is because the values of the pixels in the original inference data of digit one are zero, leading to lower pixel values in the obfuscation results. However, when more neurons are used in the first hidden layer of $O_M$, the overall darkness levels of the obfuscation results of all digits are equalized, suggesting a better obfuscation quality. The obfuscation results of $O_C$ can be found in Appendix~\ref{appendix:mnist-obf}. Similarly, we cannot interpret the obfuscation results.

\subsection{Case Study 3: American Sign Language (ASL) Recognition}
\label{asl}

In this case study, we consider an application of ASL recognition using camera-captured pictures. ASL is a set of 29 hand gestures corresponding to 26 English letters and three other special characters representing the meanings of deletion, nothing, and space delimiter. While ASL is a predominant sign language of the deaf communities in the U.S., it is also widely learned as a second language, serving as a lingua franca. Therefore, portable ASL recognition systems \cite{fang2017deepasl} are useful to the communications between ASL users and those who do not understand ASL. Porting the ASL recognition capability to smart glasses is desirable but also challenging due to smart glasses' limited compute power. Thus, remote inference is a solution for smart glass-based ASL recognition. As the hand gesture images caused by the embedded cameras can contain privacy-sensitive information (e.g., skin color, skin texture, gender, tattoo, location of the shot inferred from the picture background, etc), it is desirable to obfuscate the images. Thus, we apply ObfNet to ASL recognition.

\subsubsection{Data preparation}

We use an ASL dataset \cite{asldataset} consisting of 87,000 static hand gesture RGB images with each sized $200 \times 200$ pixels. ig.~\ref{fig:asl-origin} shows the samples corresponding to the 29 classes of the ASL alphabet. To reduce the scale of ObfNet, we down-sample the ASL images to $64 \times 64$.

\subsubsection{Architecture of InfNet}

\begin{figure}
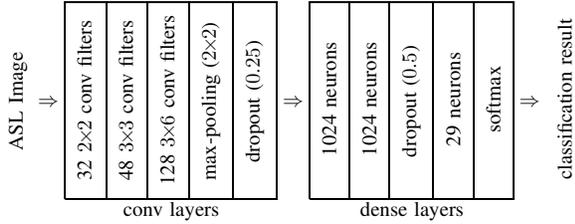

	\normalsize
	\centering
	\resizebox{0.46\textwidth}{!}{
	\begin{tabular}{ccc|c|c|c|c|c|c|c|c|c|c|c|cc}
	\cline{4-8} \cline{10-14}
	& \rotatebox[origin=c]{90}{ASL Image}  &$\!\!\!\!\Rightarrow \!\!$
	& \rotatebox[origin=c]{90}{$\;$ 32 $2\!\!\times \!\!2$ conv filters $\;$} 
	&\rotatebox[origin=c]{90}{$\;$ 48 $3\!\!\times \!\!3$ conv filters  $\;$}
	&\rotatebox[origin=c]{90}{$\;$ 128 $3\!\!\times \!\!6$ conv filters  $\;$}
	&\rotatebox[origin=c]{90}{max-pooling ($2\!\!\times \!\!2$)} 
	&\rotatebox[origin=c]{90}{dropout ($0.25$)}
	&$\!\!\Rightarrow \!\!$	 
	& \rotatebox[origin=c]{90}{1024 neurons}
	& \rotatebox[origin=c]{90}{1024 neurons}
	&\rotatebox[origin=c]{90}{dropout ($0.5$)}
	& \rotatebox[origin=c]{90}{29 neurons}
	& \rotatebox[origin=c]{90}{softmax} 
	& $\!\!\Rightarrow \!\!$  
	& \rotatebox[origin=c]{90}{classification result} 
	\\
			
	\cline{4-8} \cline{10-14}
	&\multicolumn{2}{c}{} 
	& \multicolumn{5}{c}{conv layers} 
	& \multicolumn{1}{c}{} 
	& \multicolumn{5}{c}{dense layers} 
	& \multicolumn{1}{c}{} 
	\\
	\end{tabular}
	}
	\caption{Structure of $I_C$ for ASL dataset.}
	\label{fig:inf-conv-asl}
	
\end{figure}

As ASL hand gestures have more complex patterns than the MNIST handwritten digits, we adopt a CNN-based InfNet $I_C$ Note that compared with MLP, CNN often better deals with multi-dimensional spatial data. The $I_C$ consists of three convolutional layers with 32, 64, 128 channels, a max-pooling layer, and three dense layers. We adopt adopt after the pooling layer and the second dense layer with drop rates of 0.25 and 0.5. Fig.~\ref{fig:inf-conv-asl} shows the architecture of $I_C$. %
The $I_C$ has about 111 million parameters in total.

\subsubsection{Architecture of ObfNet}

We evaluate both the MLP-based ObfNet $O_M$ and the CNN-based ObfNet $O_C$.

$O_M$ has two dense layers with ReLU activation. We vary the number of neurons in the first dense layer and evaluate how it affects the inference accuracy. $O_M$ has about 6.3 to 25.2 million parameters, depending on the number of neurons in the first dense layer.

$O_C$ consists of a convolutional layer, a pooling layer, two dense layers with ReLU activation. %
The convolutional layer filters the $64\times64\times3$ input image (i.e., $64\times 64$ RGB image) with 32 output filters of kernel size $3\times3$ and uses stride of one pixel. %
A max-pooling layer with pool size of $2\times2$ and stride of two pixels follows to reduce spatial dimensions. Two dense layers are then connected with ReLU activation. %
Two dropout layers with dropout rates of 0.25 and 0.4 are applied after the max-pooling layer and the second dense layer to prevent overfitting. %
$O_C$ has about 22 to 44 million parameters, depending on the number of neurons in the first dense layer.

\subsubsection{Inference accuracy of InfNet and ObfNet-InfNet}

The test accuracy of the trained $I_C$ is 99.82\%. This suggests that the InfNet is well trained. Multiple ObfNets are trained by following the procedure presented in \sect\ref{3.2 approach overview}.

\begin{figure}
\centering
	\includegraphics[width=.9\columnwidth]{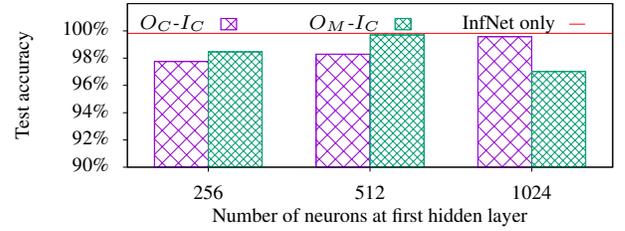}
	\caption{Test accuracy of InfNet and ObfNet-InfNet concatenations for ASL recognition.}
\label{fig:asl-acc-result}
\end{figure}

Fig.~\ref{fig:asl-acc-result} shows the test accuracy of various concatenations of ObfNets and InfNets when the number of neurons in the first hidden layer of the ObfNet varies. From Fig.~\ref{fig:asl-acc-result}, compared with the test accuracy of $I_C$, the concatenation $O_M$-$I_C$ has test accuracy drops ranging from 0.12\% to 2.81\% over various neuron number settings. When the ObfNet $O_C$ is adopted, the concatenation $O_C$-$I_C$ has test accuracy drops ranging from 1.52\% to 2.35\%. When the number of neurons in the first hidden layer increases from 512 to 1024, the test accuracy of the $O_M$-$I_C$ drops. This can be caused by overfitting, because compared with the large number of $O_M$'s parameters, the number of training samples is not large. Nevertheless, with proper configuration of the ObfNet, the smallest test accuracy drop we can achieve is 0.12\%. This shows that the ObfNet introduces little test accuracy drop for ASL recognition.

\subsubsection{Quality of obfuscation}

\begin{figure}
	\centering
	\subfigure[Original inference data]{\label{fig:asl-origin}\includegraphics[width=1\columnwidth]{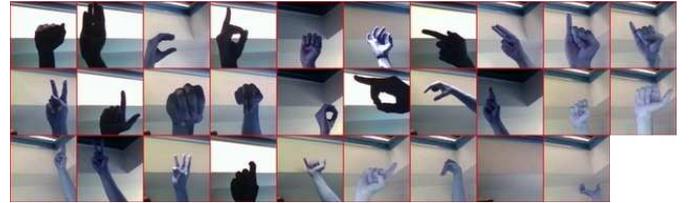}}
	\subfigure[Obfusaction results of $O_M$]{\label{fig:asl-obfnet-mlp}\includegraphics[width=1\columnwidth]{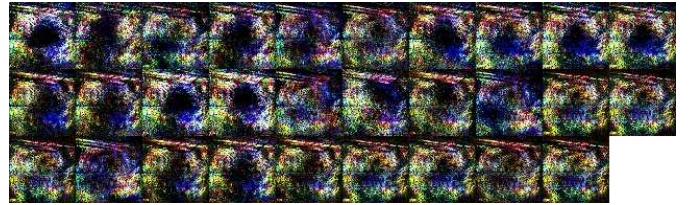}}
        \subfigure[Obfusaction results of $O_C$]{\label{fig:asl-cnn1024}\includegraphics[width=1\columnwidth]{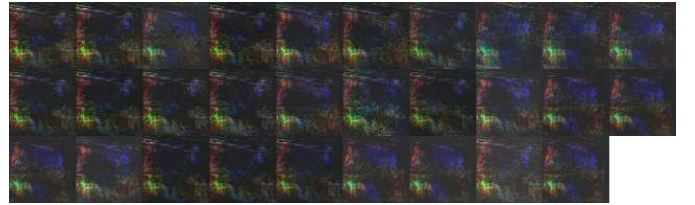}}
	\caption{Obfuscation results of ObfNet on ASL.}
	\label{fig:asl-image-result}
\end{figure}

Fig.~\ref{fig:asl-image-result} shows the visual effect of the obfuscation on the ASL samples. From Fig.~\ref{fig:asl-obfnet-mlp} and Fig.~\ref{fig:asl-cnn1024}, we cannot interpret the obfuscation results of $O_M$ and $O_C$ into any hand gestures. Note that the obfuscated samples are still RGB images. Interestingly, the obfuscation results by a certain ObfNet exhibit similar patterns. For instance, each obfuscated sample in Fig.~\ref{fig:asl-obfnet-mlp} has a dark hole in the center and a greenish circular belt around the dark hole. In fact, as the ObfNet has a large number of parameters (up to tens of million), the pattern shown in the obfuscation result is mainly determined by the ObfNet, whereas the original inference data sample with a relatively limited amount of information ($64 \times 64 \times 3 = 12288$ pixel values only) can be viewed as a perturbation.

\section{Implementation and Benchmark}
\label{sec:implementation}

This section presents the implementation of our ObfNet approach on edge/backend hardware platforms. The benchmark results on the hardware platforms give understanding on the feasibility of ObfNet in practice and interesting observations. For conciseness of presentation, we only present the results of $O_M$ trained for $I_C$ in the three case studies.

\subsection{Hardware Platforms}
\label{sub:hardware-platforms}
Our implementation uses the Coral development board \cite{coral} (referred to as Coral) and NVIDIA Jetson AGX Xavier \cite{jetson_agx} (referred to as Jetson) as the edge device and backend hardware platforms, respectively. We implement the ObfNets and InfNets of the three case study applications presented in \sect\ref{sec:case_studies} on Coral and Jetson, respectively.

Coral is a single-board computer equipped with an NXP iMX8M system-on-chip and a Google Edge TPU. Edge TPU is an inference accelerator that cannot perform ML model training. Coral sizes $8.8 \times 6\,\text{cm}^2$ and weighs about 136 grams including a thermal transfer plate and a heat dissipation fan. The power consumption of Coral is no great than $8.5\,\text{W}$. Thus, Coral is a modern edge device platform with hardware-accelerated inference capability. Note that owing to ObfNets' small-scale design, they can also run on edge devices without hardware acceleration for inference. Coral runs Mendel, a lightweight GNU/Linux distribution. We deploy the ObfNet implemented using the TensorFlow Lite library on Coral.

Jetson is a computing board equipped with a 8-core ARM CPU, 16GB LPDDR4x memory, and a 512-core Volta GPU. The GPU can accelerate DNN training and inference. Jetson sizes $10.5 \times 10.5\,\text{cm}^2$ and weighs 280 grams including a thermal transfer plate. Jetson's power rating can be configured as $10\,\text{W}$, $15\,\text{W}$, and $30\,\text{W}$. In our experiments, we configure it to run at $30\,\text{W}$ to achieve the highest compute power. Jetson can be employed as an embedded backend to serve edge devices of applications in a locality such as an office building and a factory floor. To support massive edge devices, a cloud backend can be used instead. Jetson runs Ubuntu. We deploy the InfNet implemented using TensorFlow on Jetson.

\subsection{Benchmark Results}
\label{sub:benchmark-result}
\begin{table}
  \centering
      \caption{Per-sample execution time on Coral for $O_M$ trained for $I_C$.}
      \label{tab:obf_edge}
    \begin{tabular}{|l|c|c|c|}
        \hline
        \multirow{2}{*}{Case study} & \multicolumn{3}{c|}{ObfNet execution time (ms)}                     \\ \cline{2-4}
                                         & Minimum                                    & Average & Maximum \\\hline
        FSD-$O_M$                 & 2.226                                      & 2.312   & 2.253   \\ \hline
        MNIST-$O_M$                            & 0.221                                      & 0.221   & 0.224   \\ \hline
        ASL-$O_M$                              & 11.136                                     & 11.146  & 11.170  \\ \hline
    \end{tabular}
\end{table}

\begin{table}
  \centering
      \caption{Per-sample execution time on Jetson for $I_C$.}
    \label{tab:inf_jetson}
    \begin{tabular}{|l|c|c|c|}
        \hline
        \multirow{2}{*}{Case study} & \multicolumn{3}{c|}{InfNet execution time (ms)}                     \\ \cline{2-4}
                               & Minimum                                    & Average & Maximum \\\hline
        FSD-$I_{C}$              & 0.229                                      & 0.246   & 0.289   \\ \hline
        MNIST-$I_{C}$                  & 0.158                                      & 0.174   & 0.212   \\ \hline
        ASL-$I_C$                   & 0.201                                      & 0.219   & 0.249   \\ \hline
    \end{tabular}
\end{table}

For each case study application, we measure the per-sample execution time for obfuscation on Coral and per-sample inference time on Jetson. To mitigate the uncertainties caused by the operating systems' scheduling, for each tested setting, we run ObfNet or InfNet for 100 times.

\subsubsection{Model communication overhead}
\label{sub:model-communication-overhead}
As multiple ObfNets are transmitted to the edge node for selecting, additional communication overhead is introduced. This set of measurements evaluate such communication overhead. In our implementation, the ObfNets are transmitted in the form of TensorFlow Lite FlatBuffer file (.tflite). Thus, the model communication cost can be measured by the multiplication of the number of ObfNets transmitted and the size of TensorFlow Lite FlatBuffer file of a single ObfNet. For FSD, MNIST, and ASL, a single ObfNet is 1.4 MB, 618 KB, and 1.1 MB, respectively. If a communication link throughput of 10Mbps is available, the communication of one ObfNet requires 1.12, 0.49, and 0.88 seconds for FSD, MNIST, and ASL, respectively.

\subsubsection{ObfNet and InfNet execution times and energy expenditures}
\label{sub:exec-time-and-enery}
\begin{figure*}
    \begin{minipage}{.73\textwidth}
        \centering
    \subfigure[FSD-$I_C$]
    {
        \includegraphics[width=0.28\textwidth]{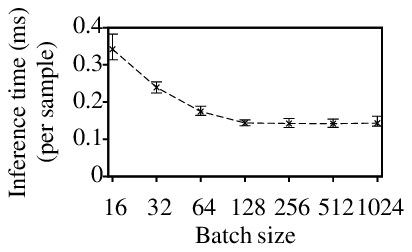}
        \label{fig:inf_time_nobn_diff_batch}
    }
    \subfigure[MNIST-$I_C$]
    {
        \includegraphics[width=0.28\textwidth]{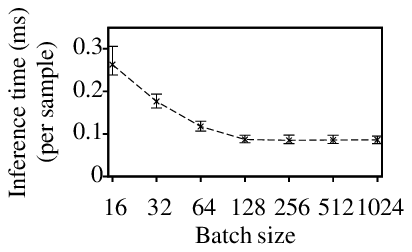}
        \label{fig:inf_time_MNIST_diff_batch}
    }
    \subfigure[ASL-$I_C$]
    {
        \includegraphics[width=0.28\textwidth]{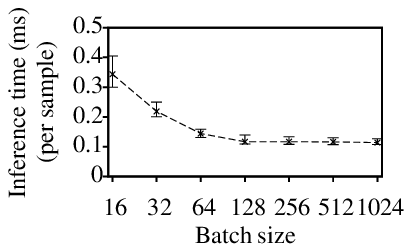}
        \label{fig:inf_time_ASL_diff_batch}
    }
    \caption{InfNet's per-sample execution time on Jetson versus batch size. Error bar represents average, maximum and minimum over 100 tests.}
    \label{fig:inf_time_diff_batch}
      \end{minipage} 
      \hspace{2em}
      \begin{minipage}{.20\textwidth}
        \centering
        \includegraphics[width=\textwidth]{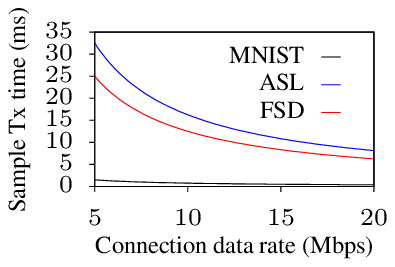}
        \caption{Data sample transmission time versus network connection data rate.}
        \label{fig:transmission_time}
      \end{minipage}
\end{figure*}

We assess the energy expenditures of executing ObfNet and InfNet as the corresponding execution times multiplied by the rated power of Coral and Jetson as mentioned in Section~\ref{sub:hardware-platforms}. Table~\ref{tab:obf_edge} shows Coral's per-sample execution times for the ObfNets designed for the three case studies when the batch size is 32. We can see that, the ObfNets need little processing time (i.e., a few milliseconds) on Coral.
Based on the average execution time per sample, the average energy expenditures of executing ObfNet on Coral are $20\,\text{mJ}$, $2\,\text{mJ}$, and $9.5\,\text{mJ}$ for FSD, MNIST, and ASL, respectively。 Table~\ref{tab:inf_jetson} shows Jetson's per-sample execution time for the InfNets designed for the three case studies when the batch size is 32. Although the InfNets have larger scales than the ObfNets, the execution times of InfNets are shorter than those of ObfNets due to Jetson's greater compute power. In TensorFlow, batch execution of inferences can improve the efficiency of utilizing the hardware acceleration. Thus, we also evaluate the impact of the batch size on the per-sample execution time of InfNets. Fig.~\ref{fig:inf_time_diff_batch} shows the results. We can see that the per-sample execution time decreases with the batch size and converges. The convergence is caused by the saturation of the hardware acceleration utilization. When the batch size is 32, the average energy expenditures of executing InfNet on Jetson are $7\,\text{mJ}$, $5\,\text{mJ}$, and $7\,\text{mJ}$ for FSD, MNIST, and ASL, respectively. From Fig.~\ref{fig:inf_time_diff_batch}, when the batch size increases, the energy expenditures of executing InfNet on Jetson can be further reduced. The above results show that the ObfNets and InfNets introduce little overhead to the edge device and the backend for the considered case study applications.

\begin{table}
  \centering
      \caption{Per-sample execution time of $I_C$ on Coral.}
    \label{tab:inf_edge}
    \begin{tabular}{|l|c|c|c|}
        \hline
        \multirow{2}{*}{Model} & \multicolumn{3}{c|}{Inference time ($ms$)}                     \\ \cline{2-4}
                               & Minimum                                    & Average & Maximum \\\hline
        FSD-$I_C$              & 13.484                                     & 14.318  & 15.137  \\ \hline
        MNIST-$I_C$                  & 7.606                                      & 8.351   & 9.095   \\ \hline
        ASL-$I_C$                    & 100.433                                    & 100.467 & 100.510 \\ \hline
    \end{tabular}
\end{table}

\subsubsection{Advantage of remote inference}
\label{sub:advantage-of-remote-inference}
Inference accelerators such as Edge TPU may enable the execution of deep InfNets on edge devices (i.e., {\em local inference}). In contrast, the remote inference scheme considered in this paper involves the transmissions of the inference data to the backend, which may incur increased latency. In this set of benchmark experiments, we put aside the need of protecting the confidentiality of InfNets as discussed in \sect\ref{sec:intro} and compare the local inference and remote inference in terms of total time delay.

Table~\ref{tab:inf_edge} shows the execution time of InfNets on Coral. Compared with the results in Table~\ref{tab:inf_jetson}, for the FSD and MNIST case study applications, the execution times on Coral are about 50x longer than those on Jetson. For ASL, it is about 480x longer. Regarding energy expenditures, although Jetson consumes 3.5 times more power than Coral, executing InfNet on it is still more power-efficient due to the much shorter execution times. The data transmission delays under the remote inference scheme are often small, because edge devices often have wideband network connections (e.g., Wi-Fi and 4G). Based on the average inference data sample sizes of the case study applications (i.e., $10\,\text{KB}$, $13\,\text{KB}$, and $0.6\,\text{KB}$ for FSD, ASL, and MNIST, respectively), Fig.~\ref{fig:transmission_time} shows the per-sample transmission times versus the network connection data rate. Analysis shows that, compared with the local inference, the remote inference achieves shorter time delays when the connection data rate is higher than $15\,\text{Mbps}$. Note that 4G connections normally provide more than $100\,\text{Mbps}$ data rate. Thus, remote inference will be more advantageous in terms of total time delay and power efficiency. 

\section{Conclusion and Future Work}
\label{sec:conclusion}

The case studies presented in this paper show that there can exist a small-scale non-linear transform in the form of a neural network, i.e., ObfNet $O(\cdot)$, such that the transformed inference data samples are mapped to the same class labels as the original inference data samples, where the mapping is the InfNet $I(\cdot)$. Formally, $\exists O(\cdot)$, $I(O(x)) = I(x)$ holds mostly, $\forall x \in \mathcal{X}$, where $\mathcal{X}$ represents the inference dataset. The evaluation also shows that the ObfNet can well protect the confidentiality of the raw form of the inference data sample $x$, through the volunteers' auditory examination on the obfuscated FSD samples and the visual examination on the obfuscated MNIST and ASL samples. Therefore, this paper presents a lightweight and unobtrusive data obfuscation approach for inference, which can be used to protect the edge devices' data privacy in the remote inference systems.

In our future work, we aim to apply the ObfNet approach for a number of heavyweight InfNets that deal with more complex auditory and visual sensing tasks such as full-fledged speech recognition and DNNs for ImageNet.


\bibliographystyle{IEEEtran}
\bibliography{reference_jiot}

\appendix
\subsection{Confusion Matrices for Other Three ObfNets in FSD Recognition}
\label{appendix: confusion-matrices}

Fig.~\ref{fig:matrix_OCIF}, Fig.~\ref{fig:matrix_OFIC}, and Fig.~\ref{fig:matrix_OFIF} show the confusion matrices for recognizing the audio inverted from the MFCC representations obfuscated by $O_C$ trained for $I_M$, $O_M$ trained for $I_C$, and $O_M$ trained for $I_M$, respectively.

\begin{figure}[p]
	\resizebox{0.49\textwidth}{!}{
		
		\begin{tabular}{l|c|c|c|c|c|c|c|c|c|c|c|c|}
			
			\multicolumn{13}{c}{\textbf{Perceived label}}\\
			\cline{3-13}
			\multicolumn{2}{c|}{}&\textbf{0}&\textbf{1}&\textbf{2}&\textbf{3}&\textbf{4}&\textbf{5}&\textbf{6}&\textbf{7}&\textbf{8}&\textbf{9}&{Accuracy}\\
			\cline{2-13}
			\multirow{10}{*}{\rotatebox[origin=c]{90}{\textbf{True label}}}
			&\textbf{0}& \yb$ $ & $1$ & $2$ & $2$ & $1$ & $1$ & $1$ & $2$ & $ $ & $ $ & $0\%$ \\
			\cline{2-13}
			&\textbf{1}& $1$ & \yb$ $ & $1$ & $2$ & $ $ & $2$ & $1$ & $1$ & $1$ & $1$ & $0\%$\\
			\cline{2-13}
			&\textbf{2}& $1$ & $1$ & \yb$2$ & $2$ & $1$ & $ $ & $ $ & $2$ & $1$ & $ $ &$20\%$\\
			\cline{2-13}
			&\textbf{3}& $ $ & $3$ & $1$ & \yb$ $ & $1$ & $1$ & $2$ & $1$ & $ $ & $1$ &$0\%$\\
			\cline{2-13}
			&\textbf{4}& $1$ & $ $ & $1$ & $1$ & \yb$2$ & $1$ & $1$ & $2$ & $1$ & $ $ &$20\%$\\
			\cline{2-13}
			&\textbf{5}& $ $ & $1$ & $2$ & $2$ & $1$ & \yb$1$ & $1$ & $1$ & $1$ & $ $ &$10\%$\\
			\cline{2-13}
			&\textbf{6}& $1$ & $1$ & $2$ & $2$ & $1$ & $1$ & \yb$ $ & $1$ & $1$ & $ $ &$0\%$\\
			\cline{2-13}
			&\textbf{7}& $ $ & $1$ & $1$ & $1$ & $1$ & $1$ & $3$ & \yb$ $ & $1$ & $1$ &$10\%$\\
			\cline{2-13}
			&\textbf{8}& $1$ & $1$ & $2$ & $1$ & $2$ & $1$ & $ $ & $ $ & \yb$ $ & $1$ &$10\%$\\
			\cline{2-13}
			&\textbf{9}& $1$ & $ $ & $1$ & $2$ & $1$ & $2$ & $1$ & $1$ & $ $ & \yb$1$ &$10\%$\\
			\cline{2-13}
			\multicolumn{13}{c}{\textbf{Overall accuracy = 7\%}}\\

		\end{tabular}
	}

	\caption{Confusion matrix for recognizing the audio inverted from the MFCC representations obfuscated by ObfNet $O_C$ that is trained for InfNet $I_M$. The matrix omits the zeros.}
	\label{fig:matrix_OCIF}
\end{figure}

\begin{figure}
	\resizebox{0.49\textwidth}{!}{
		
		\begin{tabular}{l|c|c|c|c|c|c|c|c|c|c|c|c|}
			
			\multicolumn{13}{c}{\textbf{Perceived label}}\\
			\cline{3-13}
			\multicolumn{2}{c|}{}&\textbf{0}&\textbf{1}&\textbf{2}&\textbf{3}&\textbf{4}&\textbf{5}&\textbf{6}&\textbf{7}&\textbf{8}&\textbf{9}&{Accuracy}\\
			\cline{2-13}
			\multirow{10}{*}{\rotatebox[origin=c]{90}{\textbf{True label}}}
			&\textbf{0}& \yb$ $ & $1$ & $1$ & $2$ & $1$ & $1$ & $1$ & $1$ & $2$ & $ $ & $0\%$ \\
			\cline{2-13}
			&\textbf{1}& $1$ & \yb$1$ & $2$ & $1$ & $1$ & $1$ & $2$ & $ $ & $1$ & $ $ & $10\%$\\
			\cline{2-13}
			&\textbf{2}& $1$ & $1$ & \yb$ $ & $2$ & $1$ & $1$ & $ $ & $1$ & $1$ & $2$ &$0\%$\\
			\cline{2-13}
			&\textbf{3}& $ $ & $ $ & $ $ & \yb$1$ & $3$ & $2$ & $2$ & $1$ & $ $ & $1$ &$10\%$\\
			\cline{2-13}
			&\textbf{4}& $1$ & $ $ & $2$ & $2$ & \yb$1$ & $1$ & $1$ & $1$ & $1$ & $ $ &$10\%$\\
			\cline{2-13}
			&\textbf{5}& $ $ & $1$ & $1$ & $0$ & $1$ & \yb$1$ & $3$ & $1$ & $2$ & $ $ &$10\%$\\
			\cline{2-13}
			&\textbf{6}& $1$ & $ $ & $3$ & $1$ & $ $ & $ $ & \yb$1$ & $2$ & $1$ & $1$ &$10\%$\\
			\cline{2-13}
			&\textbf{7}& $ $ & $ $ & $ $ & $4$ & $1$ & $1$ & $1$ & \yb$2$ & $1$ & $ $ &$20\%$\\
			\cline{2-13}
			&\textbf{8}& $2$ & $ $ & $2$ & $1$ & $ $ & $1$ & $1$ & $2$ & \yb$ $ & $1$ &$0\%$\\
			\cline{2-13}
			&\textbf{9}& $1$ & $ $ & $ $ & $3$ & $1$ & $1$ & $1$ & $2$ & $1$ & \yb$ $ &$0\%$\\
			\cline{2-13}
			\multicolumn{13}{c}{\textbf{Overall average = 7\%}}\\

		\end{tabular}
	}

	\caption{Confusion matrix for recognizing the audio inverted from the MFCC representations obfuscated by ObfNet $O_M$ that is trained for InfNet $I_C$. The matrix omits the zeros.}
	\label{fig:matrix_OFIC}
\end{figure}

\begin{figure}
	\resizebox{0.49\textwidth}{!}{
		
		\begin{tabular}{l|c|c|c|c|c|c|c|c|c|c|c|c|}
			
			\multicolumn{13}{c}{\textbf{Perceived label}}\\
			\cline{3-13}
			\multicolumn{2}{c|}{}&\textbf{0}&\textbf{1}&\textbf{2}&\textbf{3}&\textbf{4}&\textbf{5}&\textbf{6}&\textbf{7}&\textbf{8}&\textbf{9}&{Accuracy}\\
			\cline{2-13}
			\multirow{10}{*}{\rotatebox[origin=c]{90}{\textbf{True label}}}
			&\textbf{0}& \yb$1$ & $1$ & $3$ & $1$ & $1$ & $ $ & $2$ & $ $ & $ $ & $1$ & $10\%$ \\
			\cline{2-13}
			&\textbf{1}& $ $ & \yb$1$ & $ $ & $3$ & $1$ & $1$ & $2$ & $1$ & $1$ & $ $ & $10\%$\\
			\cline{2-13}
			&\textbf{2}& $ $ & $2$ & \yb$1$ & $1$ & $1$ & $1$ & $2$ & $ $ & $1$ & $1$ &$10\%$\\
			\cline{2-13}
			&\textbf{3}& $ $ & $1$ & $2$ & \yb$ $ & $1$ & $2$ & $1$ & $1$ & $ $ & $2$ &$0\%$\\
			\cline{2-13}
			&\textbf{4}& $2$ & $ $ & $1$ & $1$ & \yb$ $ & $1$ & $1$ & $2$ & $1$ & $1$ &$0\%$\\
			\cline{2-13}
			&\textbf{5}& $0$ & $1$ & $2$ & $3$ & $2$ & \yb$ $ & $1$ & $1$ & $$ & $ $ &$0\%$\\
			\cline{2-13}
			&\textbf{6}& $ $ & $1$ & $1$ & $3$ & $1$ & $1$ & \yb$ $ & $2$ & $1$ & $ $ &$0\%$\\
			\cline{2-13}
			&\textbf{7}& $ $ & $1$ & $1$ & $2$ & $1$ & $1$ & $2$ & \yb$1$ & $1$ & $$ &$10\%$\\
			\cline{2-13}
			&\textbf{8}& $1$ & $3$ & $3$ & $1$ & $3$ & $2$ & $ $ & $1$ & \yb$ $ & $ $ &$0\%$\\
			\cline{2-13}
			&\textbf{9}& $2$ & $ $ & $2$ & $1$ & $1$ & $2$ & $1$ & $1$ & $ $ & \yb$ $ &$0\%$\\
			\cline{2-13}
			\multicolumn{13}{c}{\textbf{Overall accuracy = 4\%}}\\

		\end{tabular}
	}

	\caption{Confusion matrix for recognizing the audio inverted from the MFCC representations obfuscated by ObfNet $O_M$ that is trained for InfNet $I_M$. The matrix omits the zeros.}
	\label{fig:matrix_OFIF}
\end{figure}

\clearpage
\subsection{Obfuscation Results of ObfNet $O_C$ on MNIST}
\label{appendix:mnist-obf}

\begin{figure}[h]
	\centering
	\subfigure[Original inference data]{\label{fig:mnist_origin}\includegraphics[width=1\columnwidth]{images/mnist/input}}
	\subfigure[Obfusaction results of $O_C$ with 8 neurons in the first dense layer]{\label{fig:mnist_cnnobf8}\includegraphics[width=1\columnwidth]{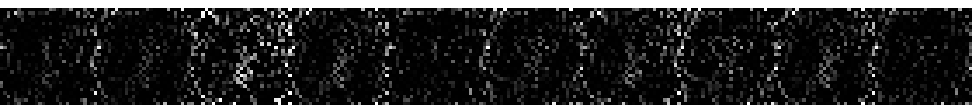}}
	\subfigure[Obfusaction results of $O_C$ with 16 neurons in the first dense layer]{\label{fig:mnist_cnnobf16}\includegraphics[width=1\columnwidth]{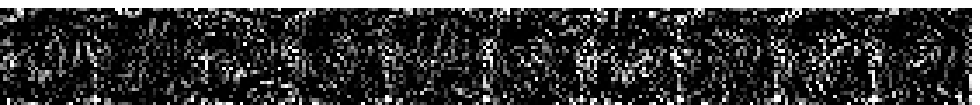}}
	\subfigure[bfusaction results of $O_C$ with 32 neurons in the first dense layer]{\label{fig:mnist_cnnobf32}\includegraphics[width=1\columnwidth]{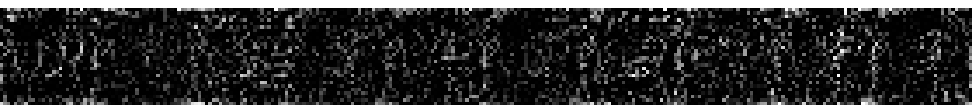}}
	\subfigure[Obfusaction results of $O_C$ with 64 neurons in the first dense layer]{\label{fig:mnist_cnnobf64}\includegraphics[width=1\columnwidth]{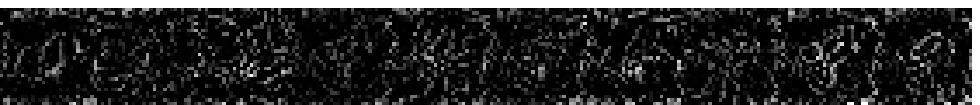}}
	\subfigure[Obfusaction results of $O_C$ with 128 neurons in the first dense layer]{\label{fig:mnist_cnnobf128}\includegraphics[width=1\columnwidth]{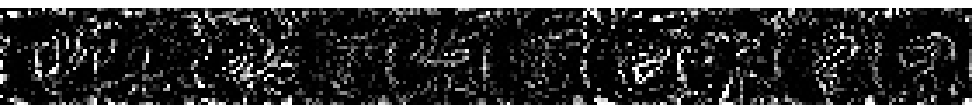}}
	\subfigure[Obfusaction results of $O_C$ with 256 neurons in the first dense layer]{\label{fig:mnist_cnnobf256}\includegraphics[width=1\columnwidth]{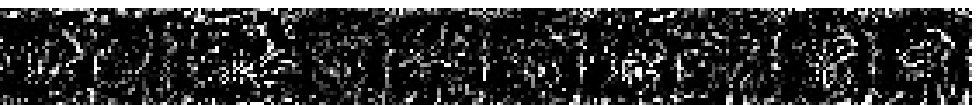}}
	\subfigure[Obfusaction results of $O_C$ with 512 neurons in the first dense layer]{\label{fig:mnist_cnnobf512}\includegraphics[width=1\columnwidth]{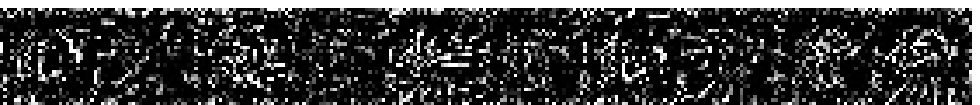}}

	\caption{Obfuscation results of ObfNet $O_C$ on MNIST.}
	\label{fig:mnist-image-result-oc}
\end{figure}

Fig.~\ref{fig:mnist-image-result-oc} shows the obfuscation results of $O_C$ on MNIST.

\begin{IEEEbiography}[{\includegraphics[width=1in,height=1.25in,clip,keepaspectratio]{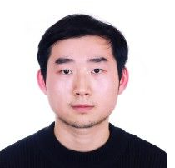}}]{Dixing Xu}
  is a fourth-year Information and Computer Science undergraduate student at Xi'an Jiaotong-Liverpool University (XJTLU), China. He is currently a visiting student researcher at Zhejiang University (ZJU), China. Previously, he was a research assist at School of Computer Science and Engineering (SCSE), Nanyang Technological University (NTU), Singapore. His research interests include machine learning system and Internet of Things (IoT).
\end{IEEEbiography}

\begin{IEEEbiography}[{\includegraphics[width=1in,height=1.25in,clip,keepaspectratio]{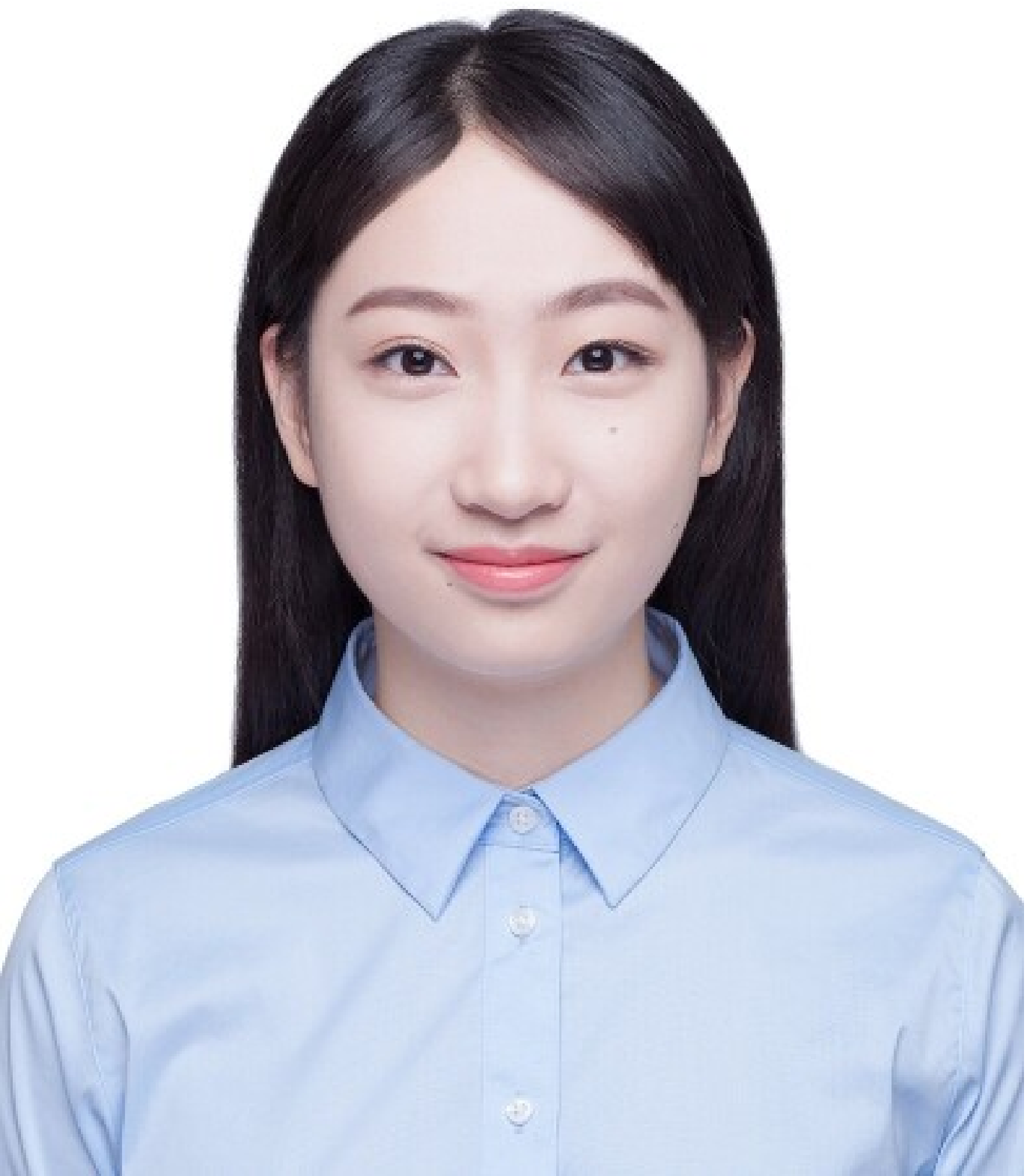}}]{Mengyao Zheng}
  is an undergraduate student (Year 3 student) at XJTLU, studying in Financial Mathematics. Previously, Mengyao was visiting SCSE of NTU. During this period, she worked with Prof. Rui Tan on privacy-preserving machine learning in IoT.
\end{IEEEbiography}

\begin{IEEEbiography}[{\includegraphics[width=1in,height=1.25in,clip,keepaspectratio]{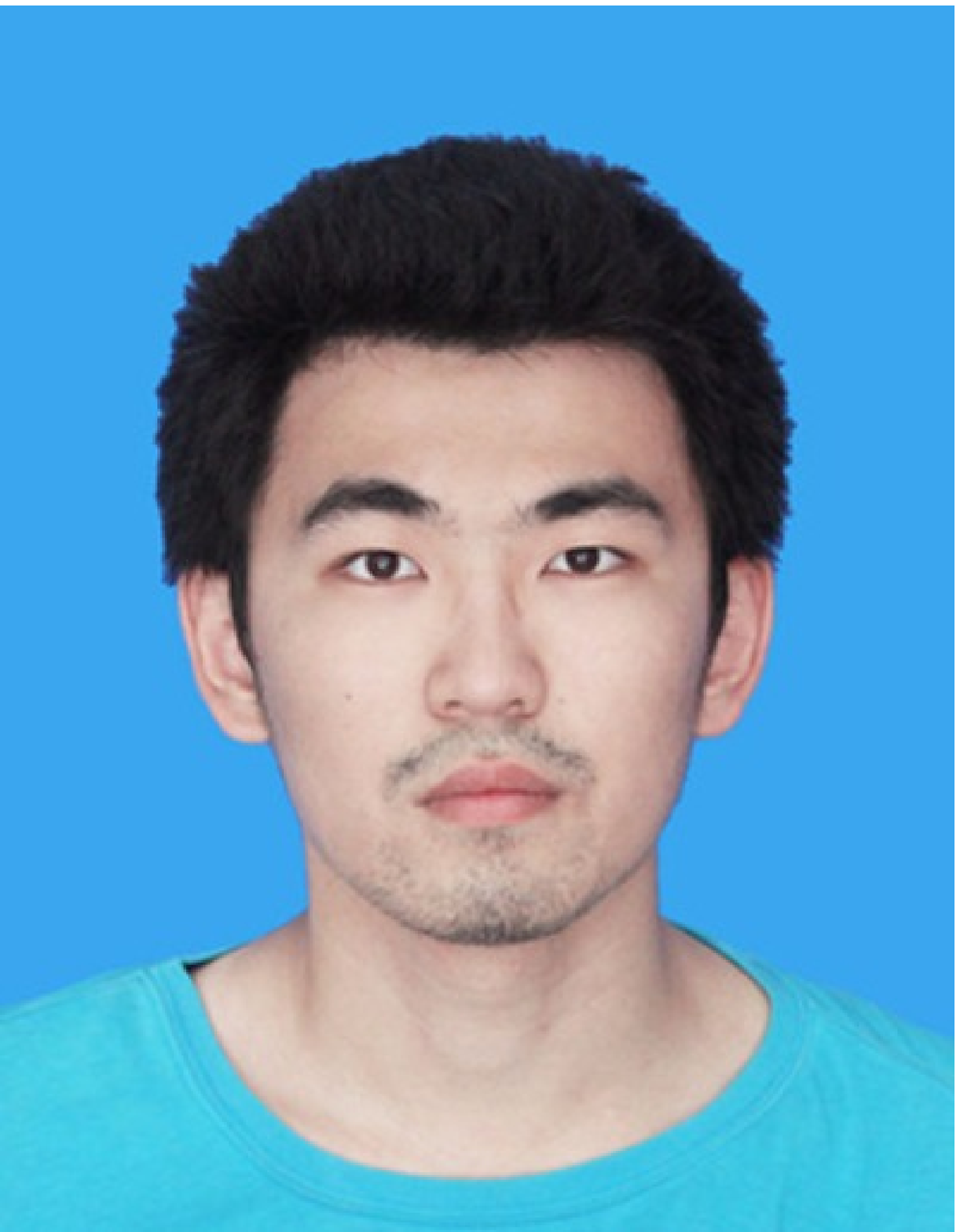}}]{Linshan Jiang}
  is a Ph.D. candidate at SCSE of NTU. He received his bachelor degree in Communication Engineering from Southern University of Science and Technology, China, in 2016. His research interests include secure and privacy in AIoT system, non-functional requirements of IoT.
\end{IEEEbiography}

\begin{IEEEbiography}[{\includegraphics[width=1in,height=1.25in,clip,keepaspectratio]{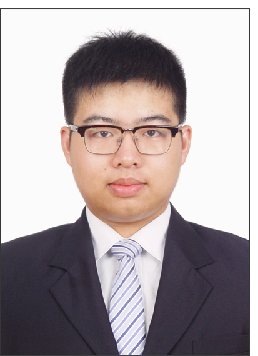}}]{Chaojie Gu}
  is a Ph.D. candidate at SCSE of NTU. He received his B.Eng. degree from Harbin Institute of Technology, China, in 2016. His research interests include IoT and Low-Power Wide Area Networks.
\end{IEEEbiography}

\begin{IEEEbiography}[{\includegraphics[width=1in,height=1.25in,clip,keepaspectratio]{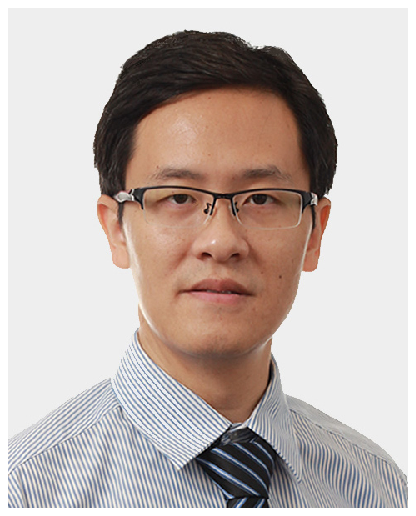}}]{Rui Tan} (M'08-SM'18) is an Assistant Professor at SCSE of NTU. Previously, he was a Research Scientist (2012-2015) and a Senior Research Scientist (2015) at Advanced Digital Sciences Center, a Singapore-based research center of University of Illinois at Urbana-Champaign (UIUC), a Principle Research Affiliate (2012-2015) at Coordinated Science Lab of UIUC, and a postdoctoral Research Associate (2010-2012) at Michigan State University. He received the Ph.D. (2010) degree in omputer science from City University of Hong Kong, the B.S. (2004) and M.S. (2007) degrees from Shanghai Jiao Tong University, China. His research interests include cyber-physical systems, sensor networks, and ubiquitous computing systems. He received the Best Paper Awards from IPSN'17, CPSR-SG'17, Best Paper Runner-Ups from IEEE PerCom'13 and IPSN'14.
\end{IEEEbiography}

\begin{IEEEbiography}[{\includegraphics[width=1in,height=1.25in,clip,keepaspectratio]{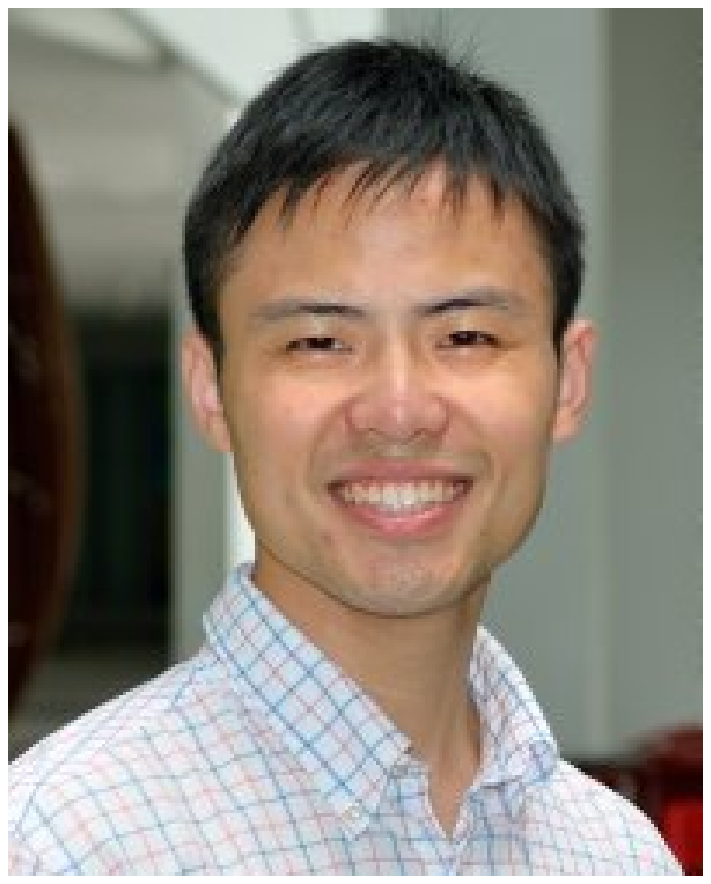}}]{Peng Cheng}
  (M'10) received the B.Sc. degree in automation and the Ph.D. degree in control science and engineering, from ZJU, in 2004 and 2009, respectively. From 2012 to 2013, he worked as Research Fellow in Information System Technology and Design Pillar, Singapore University of Technology and Design. He is currently a Professor with the College of Control Science and Engineering, ZJU. His research interests include control system security, cyber-physical systems.
\end{IEEEbiography}

\end{document}